\pdfoutput=1
\documentclass[11pt]{article}

\usepackage[]{ACL2023}

\usepackage{times}
\usepackage{latexsym}

\usepackage[T1]{fontenc}
\usepackage[utf8]{inputenc}

\usepackage{microtype}

\usepackage{inconsolata}

\usepackage{enumerate}
\usepackage{graphicx}
\usepackage{multirow}
\title{ERAGent: Enhancing Retrieval-Augmented Language Models with Improved Accuracy, Efficiency, and Personalization
}

\author{Yunxiao Shi \and Xing Zi \and Zijing Shi \and Haimin Zhang \and Qiang Wu \and Min Xu* \\
University of Technology Sydney \\
Broadway, Sydney, NSW 2007, Australia \\
\texttt{Yunxiao.Shi@student.uts.edu.au} \\
\texttt{min.xu@uts.edu.au} \\
}

\begin{document}
\maketitle
\begin{abstract}
Retrieval-augmented generation (RAG) for language models significantly improves language understanding systems. The basic retrieval-then-read pipeline of response generation has evolved into a more extended process due to the integration of various components, sometimes even forming loop structures. Despite its advancements in improving response accuracy, challenges like poor retrieval quality for complex questions that require the search of multifaceted semantic information, inefficiencies in knowledge re-retrieval during long-term serving, and lack of personalized responses persist. Motivated by transcending these limitations, we introduce ERAGent, a cutting-edge framework that embodies an advancement in the RAG area. Our contribution is the introduction of the synergistically operated module: Enhanced Question Rewriter and Knowledge Filter, for better retrieval quality. Retrieval Trigger is incorporated to curtail extraneous external knowledge retrieval without sacrificing response quality. ERAGent also personalizes responses by incorporating a learned user profile. The efficiency and personalization characteristics of ERAGent are supported by the Experiential Learner module which makes the AI assistant being capable of expanding its knowledge and modeling user profile incrementally. Rigorous evaluations across six datasets and three question-answering tasks prove ERAGent's superior accuracy, efficiency, and personalization, emphasizing its potential to advance the RAG field and its applicability in practical systems.

\end{abstract}

\section{Introduction}

Large Language Models (LLMs) represent a significant leap in artificial intelligence, with breakthroughs in generalization and adaptability across diverse tasks \cite{zero_shot_llm, PaLM}. 
However, challenges such as hallucinations \cite{ReAct}, temporal misalignments \cite{temporal_adaptation}, context processing issues \cite{L_Eval}, and fine-tuning inefficiencies \cite{shortcut} have raised significant concerns about their reliability. In response, recent research has focused on enhancing LLMs' capabilities by integrating them with external knowledge sources through retrieval-augmented generation (RAG) \cite{improve_by_retrieve, RAG, Atlas, DPR}. This approach significantly improves their ability to answer complex questions more accurately and contextually.

The basic RAG architecture comprises a retriever module and a read module, forming the retrieve-then-read pipeline \cite{RAG, DPR, Atlas}. However, this framework faces challenges in low retrieval quality, and may generate unreliable answers. To transcend the shortcomings, more advanced RAG modules have been developed and integrated into the basic framework. For example, the question rewriter is incorporated prior to the retrieval process, culminating in the Rewrite-Retrieve-Read pipeline \cite{Query_Rewriting, RETA_LLM}. This enhancement more precisely aligns user's question with the knowledge database, thereby boosting retrieval quality. Furthermore, models like RETA-LLM \cite{RETA_LLM} and RARR \cite{RARR} integrate a post-reading fact-checking component to further solidify the accuracy and reliability of the responses. Additional auxiliary modules such as the query router \cite{query_router} and the resource ranker \footnote{https://txt.cohere.com/rerank/} \cite{LLMLingua} have been assimilated into the RAG's architecture, offering increased versatility and customization. 
The integration of various components has transformed the RAG framework into a highly flexible system, leading to the emergence of a modular RAG paradigm \cite{modular_rag_survey}. The progression has evolved from the basic retrieval-read pipeline into a more complex pipeline featuring several linearly combined components. The development of ITER-RETGEN \cite{ITER-RETGEN} and TOC \cite{TOC} further extends to pipelines with loop structures. Despite these advancements, they come at the expense of increased time and cost due to the lengthy processes and the inherent complexities of iterative/recursive workflows.

RAG technologies have significantly propelled the development of LLMs across various domains. Notable implementations such as ChatENT \cite{ChatENT}, Clinfo.ai \cite{ClinfoAI}, Almanac \cite{Almanac}, and LLM-AMT \cite{LLM_AMT} have made profound impacts in AI-based assistants for healthcare, and platform Consensus \footnote{https://consensus.app/} facilitate scientific researches in education area. Similarly, RAG-based agent for auto trading in finance \cite{trading_agent} and explaining legal concepts in law  \cite{RAG_law_concepts} also demonstrate their versatility and effectiveness in addressing complex, sector-specific inquiries. 

Despite significant advancements in RAG framework and its broad application, the current RAG technique still grapples with certain unresolved and frequently overlooked deficiencies. 

\noindent \textbullet\ \textbf{Deficiency in the question rewriter}. This module aims to clarify user's question for more effective information retrieval \cite{Query_Rewriting, RETA_LLM}, falls short in scenarios involving overly granular question encompassing multifaceted information. This issue highlights the need for the module's evolution towards reformulating intricate questions into broader, more search-friendly queries. 

\noindent \textbullet\ \textbf{Redundant retrieval undermines response efficiency.} A critical inherent limitation of the current RAG systems
lies in their inability to retain previously retrieved knowledge when utilizing LLMs as black-box components. This shortcoming critically hampers their effectiveness in long-term serving, a scenario where applying previous retrieved knowledge is prevalently happened. Consequently, the RAG systems often redundantly process questions similar to those encountered in historical conversations.

\noindent \textbullet\ \textbf{Challenge in determining retrieval timing.} RAG-based AI assistant encounter challenges in determining when to engage external knowledge retrieval. Indiscriminate retrieval, particularly when it is unnecessary, significantly compromises the efficiency of the RAG system.

\noindent \textbullet\ \textbf{Uniform response for different users.} Current research on RAG primarily aims to enhance the factual accuracy of the generated responses. However, in real-world applications tailored to vary users with diverse preferences, the alignment of response with user's personalization is equally critical. Studies currently fall short in providing individual response services.

To address existing challenges, we propose the Enhanced RAG Agent (ERAGent), incorporating significant upgrades for improved functionality. Key improvements of ERAGent include an Enhanced Question Rewriter that generates clarified questions as well as fine-grained queries for richer knowledge retrieval. A Retrieval Trigger module employs a model-based judgment to selectively engage knowledge retrieval, thereby improving response efficiency. A Knowledge Filter module leverages NLI to sieve through retrieved information for relevance. A standout feature is the Personalized LLM Reader, tailoring responses to individual user contexts. Moreover, the Experiential Learner module enables ERAGent to learn user's profile from previous interactions, supporting the Retrieval Trigger and the Personalized LLM Reader, respectively. ERAGent provides a cutting-edge prototype for the development of RAG-based AI assistants, ensuring their ease of application and effectiveness in practical scenarios.
Through comprehensive experiments across three distinct question-answering tasks and six diverse datasets, we have demonstrated the superior performance of our ERAGent framework. The results not only underscore ERAGent's robust effectiveness but also elucidates the synergistic interplay among its integral components, further highlighting its potential to revolutionize the field with its accurate, efficient, and user-centric design.

In summary, our ERAGent framework integrates cutting-edge components and technologies—including Enhanced Question Rewriting, Retrieval Efficiency Optimization, an Innovative Knowledge Filter, Personalized Responses, and Pioneering Experiential Learning. Comprehensive evaluations across varied datasets and question-answering tasks confirm ERAGent's superior performance, robustness, precision, efficiency, learnability, and personalization.

\section{Related Work}
\subsection{Retrieval Augmented Generation}
Retrieval Augmented Generation (RAG) \cite{RAG} leverages a retriever that provides substantial external information for improving the generated output of LLMs. This strategy utilizes knowledge in a parameter-free manner, circumvents the high training costs of LLMs' parameterized knowledge. Furthermore, it alleviates the hallucination issues in LLMs, significantly enhancing the factual accuracy and relevance of the generated content. The concept of RAG is rooted in the DrQA framework \cite{DrQA}, which marked the initial phase of integrating retrieval mechanisms with Language Models (LMs) through heuristic retrievers like TF-IDF for sourcing evidence. Subsequently, RAG underwent evolution with the introduction of Dense Passage Retrieval \cite{DPR}, and further advancements in RAG \cite{RAG} and REALM \cite{IC-RALM}. These methods utilize pre-trained transformers and are characterized by the joint optimization of retrieval and generation components. Recent advancements have extended RAG's capabilities by integrating Large Language Models (LLMs). Exemplary developments such as REPLUG \cite{REPLUG} and IC-RALM \cite{IC-RALM} demonstrate the potent generalization abilities of LLMs in zero-shot or few-shot scenarios. These models are capable of following complex instructions, understanding retrieved information, and utilizing limited demonstrations to generate high-quality responses. 

\subsection{RAG Frameworks}
The basic framework of RAG is employs a dual-module structure consisting of a retrieval module and a reading module. Despite its demonstrable prowess in improving response generation greatly in relevance, it is still deficient in answering ambiguous question and generating accurate answer when applied to complex real-world applications. 

To transcend these limitations, a spate of endeavors has sought to refine and expand upon this dual-step pipeline. Notably, the introduction of a question rewriter within the Rewrite-Retrieve-Read framework \cite{Query_Rewriting} aligns user's question with the corresponding knowledge base to enhance the relevance of information retrieval. Furthermore, the RETA-LLM tooklit \cite{RETA_LLM} further incorporates a fact-checking module to avoid factual inaccuracies in the generated answers, thereby establishing a Rewrite-Retrieve-Read-Checking pipeline. Expanding from a singular pipeline framework to a multifaceted pipeline, the Semantic Router \footnote{\url{https://github.com/aurelio-labs/semantic-router}} incorporates a Routing module, enables the dynamic allocation of questions to designated flows, thus facilitating a tailored solutions to various types of questions.

The pipelines articulated within the aforementioned RAG Framework are predominantly linear and devoid of cyclical loops. Recent innovations also includes pipelines featuring reciprocal pipeline. For example, ITER-RETGEN \cite{ITER-RETGEN} employs a strategy where the response generated by retrieval-read is used as context to iteratively retrieve again, improving the quality of retrieval and thereby gradually enhancing the quality and relevance of the generated answer. Moreover, TOC \cite{TOC}, further explore the potential of pipeline structured in a recursive form. However, RAG with cyclical and recursive pipeline often precipitate inefficiencies, escalating costs, challenges in parallelization, and exceed the input token limitation of LLMs, which collectively undermine their practical applicability.

In contrast to the current RAG frameworks, our ERAGent is positioned as an more advanced and comprehensive solution, structured in a linear process to provide accurate, efficient, and individual responses. Contingent upon the practical applicability and requirements, our framework and components are designed to be flexible, enabling modifications or integration within the frameworks of pipelines that are more intricate and possess retrospective, ring, or cyclic structures. Thereby potentially enhancing the efficiency of these complex processes.

\subsection{RAG Modules}
The RAG framework facilitates the integration of various functional modules, which play pivotal roles in enhancing the modularity and adaptability of the system. The current research has introduced several innovative modular components, which can be categorized as follows:

\textit{Question Rewriter:} A notable advancement within the Rewriter-Retrieve-Read Framework \cite{Query_Rewriting} involves the employment of LLMs, with or without the augmentation of reinforcement learning strategies, to rephrase questions.

\textit{Question Router:} The Semantic Router exemplifies this category by analyzing the semantic essence of questions to allocate them appropriately among distinct RAG pipelines.

\textit{Knowledge Retriever:} This module excels in retrieving semantically relevant text segments from dense textual corpora through Embedding Retrieval techniques. Furthermore, API Search Engine Retrieval is a practical approach in scenarios lacking a localized in-domain database or when there is a necessity to utilize real-time online resources. Additionally, NL2SQL Retrieval, as detailed in \cite{zeroshot_NL2SQL}, specializes in navigating tabular data to unearth statistical insights.

\textit{LLM Reader:} The Reader module harnesses the capabilities of LLMs to comprehend the retrieved knowledge, subsequently formulating responses to posed questions.

\textit{Fact Checking/Revising:} Innovations such as RETA-LLM \cite{RETA_LLM} and RARR \cite{RARR} looply re-engage the retrieval-read process to verify or amend the outputs of LLMs.

\begin{figure*}[t]
  \centering
    \includegraphics[width=6.2 in]{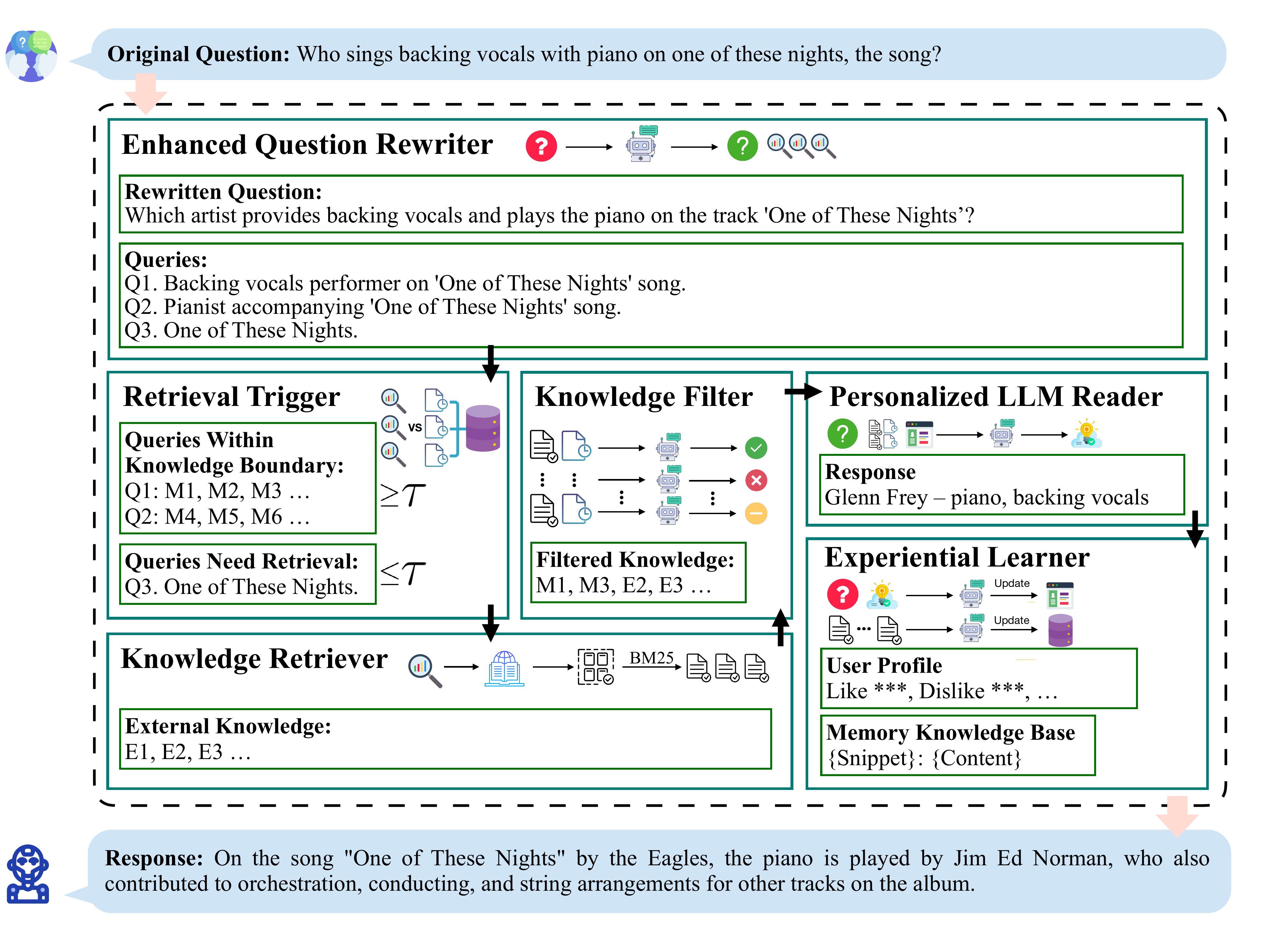}
    \vspace{-4mm}
   \caption{The ERAGent Framework.}
    \label{fig: framework}
\end{figure*}

\section{Methodology}
In this part, we firstly delve into the intricacies of our proposed pioneering framework ERAGent, and elucidate the details for each component.

\subsection{Framework Pipeline} \label{sec: framework}
The ERAGent framework is composed of the following integral components, each crafted to fulfill distinct yet synergistic functions:

\textbf{Enhanced Question Rewriter:} 
This module not only refines users' original questions, often colloquial or jargon-laden, into semantically enriched, standardized queries, but also creates fine-grained queries targeting specific semantic aspects to improve retrieval and response quality in subsequent knowledge retrieval stages.

\textbf{Retrieval Trigger:} The module examines the historical conversation experiences between
the AI assistant and the user, and assesses the knowledge boundaries. It assesses whether the query exceeds the scope of the current knowledge. For queries that exceed the knowledge boundary, the trigger is activated, necessitating the supplementation of external knowledge. 

\textbf{Knowledge Retriever:} This module is tasked with retrieving external knowledge that contributes to answering the user's question.

\textbf{Knowledge Filter:} As a post-retrieval component, this module refine the retrieved knowledge by filtering out irrelevant context to prevent the adverse effects of extraneous information on accurately answering the question.

\textbf{Personalized LLM Reader:} This module integrates the rewritten question, filtered knowledge, and description about user's preferences into a prompt, utilizing the in-context learning capabilities of LLM to generate individual answer to the question.

\textbf{Experiential Learner:} This component empowers the AI assistant to engage in a continuous learning process from historical interactions by establishing a knowledge database and maintaining a text description. This database is dedicated to preserving knowledge acquired over time and the text description continuously summarize and update the user's preferences and profile information throughout the service delivery process.

The ERAGent framework and pipeline is illustrated in Figure~\ref{fig: framework}. It starts with the user putting forward the Original Question and then the Enhanced Question Rewriter module rewrites the ambiguous question into an explicit Rewritten Question, while the module generates multiple Queries for later retrieval of knowledge. Each Query is evaluated by the Retrieval Trigger as to whether it is within or outside the Knowledge Boundary of the AI assistant. 
For Queries requiring information retrieval, the Knowledge Retriever module identifies relevant documents. In cases where Queries fall within the scope of the AI assistant's Knowledge Boundary, the relevant information is sourced from the Memory Knowledge Base. This memory information, along with externally retrieved documents, is then collectively assessed for relevance by the Knowledge Filter module. The knowledge that remains after this assessment is termed Filtered Knowledge. The Filtered Knowledge, combined with the Rewritten Question and the User Profile, informs the individual response generation by the Personalized LLM Reader. Both the User Profile and the Memory Knowledge Base are dynamically updated by the Experiential Learner module throughout the serving process of the AI assistant.

\begin{figure}[t]
  \centering
    \includegraphics[width=3 in]{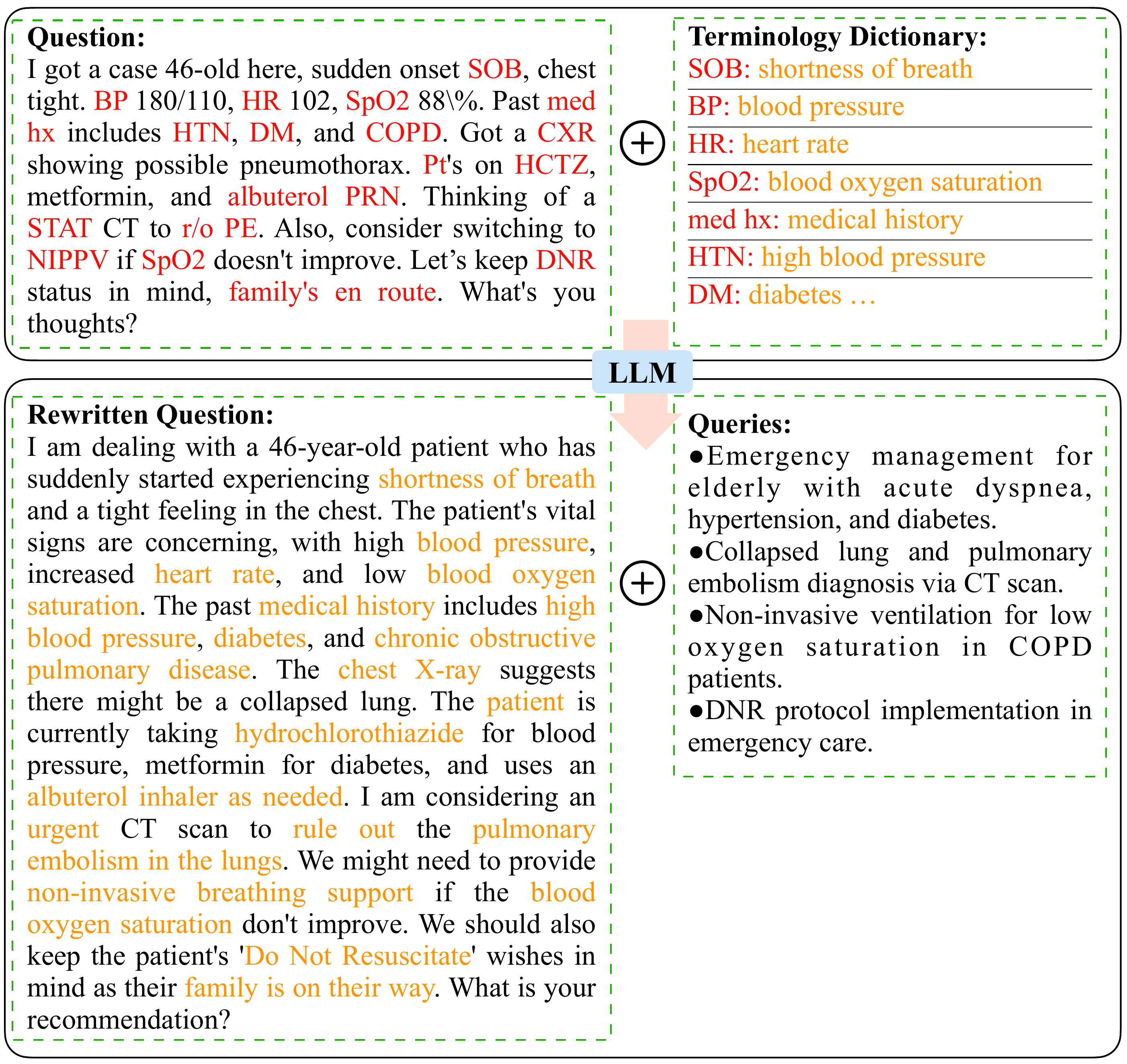}
    \vspace{-7mm}
   \caption{A case study on the application of the Enhanced Question Rewriter module in the clinical medicine area.}
    \label{fig: query_rewriter}
\end{figure}

\subsection{Enhanced Question Rewriter}
The pivotal task of this module involves the semantic augmentation of the Original Question into a Rewritten Question, as well as the generation of proper Queries underpinned by a nuanced understanding of the question. As illustrated in Figure~\ref{fig: query_rewriter}, within complex clinical scenarios, a doctor ask an AI agent for advice in a highly colloquial and jargon-filled manner. 

An optional preparatory work is to construct a comprehensive terminology dictionary, developed through extensive research and expert collaboration, encompassing a wide terms, abbreviations, and jargon specific to the specific domain. Then the terminology dictionary will be combined with the Original Question for prompting LLM to discern and substitute colloquial expressions with standard terminologies. 

This process harnesses the close-source LLM accessible via API, such as GPT-3.5. Alternatively, it is also applicable to fine-tune a small-scale LLM like Falcon 1B \cite{Falcon} with LoRa \cite{lora} finetuning.

\subsection{Retrieval Trigger}
Given the importance and practicality of calibration-based judgment methods, we have adopted "popularity" as a metric to estimate an AI assistant's mastery over a specific knowledge.

Let $M = \{M_1, M_2, \ldots, M_n\}$ be the set of knowledge snippets from the Memory Knowledge Base and let $q_i$ denote one of the generated Queries. The similarity between query $q_i$ and one of Memory Knowledge snippet $M_j$ is denoted by $S(q_i, M_j)$. A similarity threshold, $\tau$, is set to 0.6 to evaluate the relevance of Memory Knowledge instances to the query.

The popularity of query $q_i$, denoted by $Pop(q_i)$, is defined as:
$$Pop(q_i) = \left| \{M_j \in M \mid S(q_i, M_j) \geq \tau \} \right|$$
where $| \cdot |$ indicates the cardinality of the set, i.e., the number of Memory Knowledge instances $M_j$ that have a similarity score with $q_i$ greater than or equal to $\tau$.

To determine whether a query $q_i$ lies within or outside the Knowledge Boundary based on its popularity, we use a threshold $\theta$, the default is set at 3. Therefore, the conditions for a query being inside or outside the Knowledge Boundary can be defined as follows:

A query $q_i$ is considered inside the Knowledge Boundary if:
$$ Pop(q_i) \geq \theta $$

Conversely, a query $q_i$ is considered outside the Knowledge Boundary if:
$$ Pop(q_i) < \theta $$

This formulation allows for a precise and quantitative assessment of an AI assistant's knowledge pertaining to a specific query $q_i$, based on the popularity metric and the established thresholds $\tau$ and $\theta$.

\subsection{Knowledge Retriever}
A plethora of sophisticated search and retrieval algorithms can be adeptly applied in this module. Specifically, ERAGent leverages API Search Engine Retrieval technology (e.g., Bing Search v7) to exploit current online resources. The methodology involves selecting the top-ranking items from the API search results, subsequently accessing the website content of each searched item. Through the application of the BM25 algorithm, texts of high relevance are extracted, serving as the knowledge retrieved for latter use.

\subsection{Knowledge Filter}

Although the retrieved information have already been sorted by the search engine and then text chunks with low semantic similarity are filtered out. However, these documents, which exhibit a high degree of semantic similarity, discuss content that is relevant yet ambiguous in relation to the question at hand, in fact do not contain the accurate information necessary to correctly answer the question, thereby obscuring the judgement of LLMs. Therefore, we focus on ensuring the pertinence of the knowledge that can real inform subsequent accurate response generation. This component refines the corpus of retrieved knowledge, synergizing it with the Rewritten Question to produce an optimized information subset, termed Filtered Knowledge.

Instead of directly asking the LLM whether a knowledge instance is relevant to answering the question, we leverage LLM to perform the NLI task for filtering out irrelevant context. Specifically, for a given question and the corresponding knowledge context, an NLI task is to determine if the context (premise) substantiates a question-answer duo (hypothesis). We retain the context only if it is classified as 'entailment'. Conversely, if all the contexts are categorized as either 'contradiction' or 'neutral', the filter will designate the knowledge as empty. Then the subsequent response generation will proceed without external knowledge augmentation, embodying a "back-off strategy." 

\subsection{Personalized LLM Reader}
The function of LLM Reader is to generate an accurate response. Beyond merely integrating Filtered Knowledge with the Rewritten Question to formulate prompts, our enhancement involves the incorporation of the User Profile. This unique prompt enables the LLM's responses is specially generated tailoring for the user's preferences.

\subsection{Experiential Learner} \label{sec: Experiential_Learner}
The Experiential Learner module is designed with the objective of expanding the boundaries of knowledge by learning from historical dialogues between the user and the AI assistant. This endeavor aims to enhance the efficiency of responding to queries by reducing the time and cost associated with the Knowledge Retriever module in subsequent dialogues. Concurrently, the integration of highly relevant historical knowledge significantly improves the quality of knowledge, which is conducive to generating high-quality responses. Furthermore, this module also dedicates itself to learning about the user's topic interests, preferences, and question demands by analyzing historical conversations. By summarizing personal information gleaned from these interactions and incorporating it into the prompts of the LLM Reader module, the module facilitates the personalization of responses, thereby elevating the quality of responses from a personalization standpoint.

In this module, we have devised a Memory Knowledge Database intended to catalog the non-parametric knowledge continuously received by the AI assistant throughout its service. This knowledge is stored within the database in the form of snippet-content pairs; the snippet serves as a succinct summary description of the knowledge, whereas the content provides the specific context and detailed information pertaining to that knowledge. The update process of the Memory Knowledge Database is conducted on a round-by-round basis, meaning that the incremental knowledge involved in each question-answer cycle is systematically incorporated into the existing knowledge base.

Moreover, a User Profile has been designed to record the personalized information of the users being served. We employ LLMs for the in-depth analysis of session records between the user and the AI assistant, with the aim of constructing a nuanced and multi-dimensional User Profile. This profile is structured around four main aspects:

1. Theme Preferences: This facet scrutinizes chat records to extract thematic patterns and evaluate the user's disposition towards various topics. Attitudes are classified into three categories: Interest (demonstrating extensive engagement with the subject), Disinterest (exhibiting lack of interest or dissatisfaction), and Neutrality (showing indifferent or non-committal responses). For instance, an individual might exhibit interest in Machine Learning, disinterest in Law, and neutrality towards local cuisine.

2. Question Demands: This dimension involves the analysis of the user's inquiries to discern the underlying problem-solving intent and the characteristics of information-seeking behavior. Examples include difficulties encountered in assignments for a Film Art Analysis course or a plateau in weight loss efforts over the recent fortnight.

3. Basic Information: This part focuses on extracting basic information about the user, such as employment, place of residence, age, gender, etc.

4. Personlized Information: This part is dedicated to mining unique, personalized tags pertaining to the user, for instance, identifying individuals as vegetarians, late sleepers, coffee enthusiasts, and so forth.

Upon the conclusion of each session, the record undergoes a meticulous analysis, culminating in the integration of the incremental User Profile with the pre-existing one for updates. This dynamic process ensures that the User Profile remains ever-evolving, accurately reflecting the user's changing preferences, inquiries, and personal attributes.

\section{One-Round Open-Domain QA Task}
\label{sec: one_round_open_domain_qa}
In this section, we assess the effectiveness of the ERAGent in the one-round open-domain question-answering (QA) task. This task focuses on how accurately questions can be answered in a single round conversation. Therefore, we excise components Retrieval Trigger, and Experiential Learner from the ERAGent framework. Instead, we kept the Knowledge Filter and the Enhanced Question Rewriter modules to explore how they, individually and together, improve answer accuracy. 

\textbf{Research Questions.} We formulated our research questions as follows: (1) Does the Enhanced Question Rewriter module outperforms traditional Question Rewriter? (2) Can Knowledge Filter module effectively sift through irrelevant texts, thereby improve the response quality? 

\subsection{Experimental Settings}
\textbf{Datasets.} Three one-round open-domain QA datasets of increasing difficulty are used for evaluation. (i) The popular Natural Questions (NQ) dataset \cite{NQ}. (ii) PopQA \cite{when_not_trust_llm} is an open-domain long-tail distributions QA dataset, it was specifically designed to cover lower-popularity knowledge topics within Wikidata. 
(iii) AmbigNQ \cite{AmbigQA} provides a disambiguated version of Natural Questions (NQ) \cite{NQ}. For ambiguous questions in NQ, minimal constraints are added to break it into several similar but specific questions. Due to the API cost for recalling GPT-3.5, from each dataset, a random selection of 200 questions was extracted from test set.

\textbf{Baselines \& Evaluation Method.} The following settings are implemented to evaluate and support our methods. (i) Standard: This baseline involves the traditional retrieval-augmented approach with a basic retrieve-then-read pipeline. (ii) Rewriter: Building upon the Standard setting, a traditional Question Rewriter module is introduced prior to the retrieval process. (iii) Rewriter+: Building upon the Standard setting, an Enhanced Question Rewriter module is introduced prior to the retrieval process.(iv) Filter: Augmenting the Standard setting, a Knowledge Filter module is incorporated before the LLM Reader. (v) Rewriter+\&Filter: This setting combines the Enhance Question Rewriter module with the Knowledge Filter module on the basis of Standard setting. 

We adopt the metric Exact Match (EM), Precision, Recall, and the Hit Rate for evaluating the response accuracy. We utilize GPT-3.5, setting the temperature parameter to 0, to perform tasks including question rewriting, query generation, question answering, and NLI task. Detailed information on the design of prompts is available in Appendix~\ref{sec:appendix_a}.

\begin{table*}[t]
\caption{Evaluation Metrics for response accuracy in One-Round Open-Domain Question Answering.}
\label{tab: one-round open-domain}
\centering
\begin{tabular}{cccccc} 
\hline
Method          & Dataset                  & EM             & Precision      & Recall         & Hit Rate        \\ 
\hline
Standard        & \multirow{5}{*}{NQ}      & 38.00          & 54.21          & \textbf{77.38} & 55.00           \\
Rewriter        &                          & 35.50          & 55.16          & 73.48          & 54.00           \\
Rewriter+       &                          & 38.00          & \textbf{57.24} & 73.93          & 55.50           \\
Filter          &                          & 36.00          & 53.82          & 76.89          & 52.50           \\
Rewriter+Filter &                          & \textbf{40.50} & 56.84          & 75.50          & \textbf{58.50}  \\ 
\hline
Standard        & \multirow{5}{*}{PopQA}   & 29.50          & 68.43          & 44.38          & 35.00           \\
Rewriter        &                          & 30.00          & 64.46          & 46.50          & 35.00           \\
Rewriter+       &                          & 32.00          & 69.54          & 47.59          & 37.50           \\
Filter          &                          & 34.00          & \textbf{70.17} & 46.17          & 38.00           \\
Rewriter+Filter &                          & \textbf{36.00} & 69.13          & \textbf{48.40} & \textbf{40.50}  \\ 
\hline
Standard        & \multirow{5}{*}{AmbigNQ} & 38.00          & 62.03          & 64.45          & 52.00           \\
Rewriter        &                          & 41.00          & 65.35          & 63.78          & 55.50           \\
Rewriter+       &                          & 45.50          & 67.70          & 65.84          & 58.50           \\
Filter          &                          & 39.50          & 65.18          & 64.88          & 55.00           \\
Rewriter+Filter &                          & \textbf{47.00} & \textbf{69.82} & \textbf{67.06} & \textbf{63.50}  \\
\hline
\end{tabular}
\vspace{-7pt}
\end{table*}

\subsection{Results}
Experimental results on one-round open-domain QA task are reported in Table~\ref{tab: one-round open-domain}. 

\textbf{The effect of Knowledge Filter.} Comparing the Filter setting to the Standard setting, the Filter manifests enhancement over the Standard on the PopQA and AmbigNQ datasets, while a decrement is observed on the NQ dataset. This phenomenon aligns with findings reported in \cite{LLM_filter}, attributed to the NLI-based filtering method's overly stringent criteria, which inadvertently discards potentially relevant knowledge that, despite not being classified as entailment, could contribute to generating accurate responses.

\textbf{The effect of the Enhanced Question Rewriter.} The result of the Rewriter+ setting reveals improvements in all datasets and metrics, except for the decrease of Recall on NQ dataset, compared to the Standard setting. A comparative between Rewriter+ and Rewriter on all datasets showed performance improvement, and this improvement becomes apparent as the dataset becomes difficult. This underscores the efficacy of the proposed Enhanced Question Rewriter module in scenarios requiring fine-grained, multi-faceted search for ambiguous queries.

\textbf{The joint effect of Knowledge Filter and Enhance Question Rewriter.} 
The Rewriter+\&Filter setting demonstrates notable enhancements in performance across all datasets and metrics when juxtaposed with the deployment of a single module in isolation. This underscores that the improved retrieval quality brought by comprehensive and diverse external knowledge retrieved by the Enhanced Question Rewriter can be further improved by eliminating irrelevant context through the Knowledge Filter module. As a result, the combined effect of these two modules significantly exceeds that of other all settings in terms of response quality.

\subsection{Analysis}
The Enhanced Question Rewriter module demonstrates improved performance over the basic Question Rewriter, effectively handling ambiguous and complex questions with greater precision. This superior performance is attributed to the complete identification and query of every semantic aspect of a problem. Consequently, this meticulous approach to question rewriting markedly improves the accuracy of responses.

Furthermore, the Knowledge Filter module adds another layer of refinement to the process. By sifting through the plethora of information retrieved, it meticulously identifies and discards texts that are irrelevant. This filtration process ensures that the information fed into the language model is of the highest relevance and quality, thereby significantly enhancing the accuracy of the generated responses. Its utility shines brightest in contexts where the retrieval phase yields extensive information. However, it's important to note that the efficacy of the solely utilization of the Knowledge Filter does not uniformly ensure enhanced accuracy in responses across various scenarios.

The interplay between the Enhanced Question Rewriter and the Knowledge Filter modules demonstrates a synergistic effect that is greater than their isolated application. When deployed in tandem, these modules collaboratively refine the pre-retrieval and post-retrieval processes, creating a conducive environment for augmenting LLM to generate responses of unparalleled accuracy and relevance.

\begin{table*}[t]
\caption{Evaluation Metrics for response accuracy in One-Round Multi-Hop Question Answering.}
\label{tab: one-round multi-hop reasoning}
\centering
\begin{tabular}{cccllc} 
\hline
Method                          & Dataset                   & EM             & Precision      & Recall         & Hit Rate        \\ 
\hline
Standard                        & \multirow{4}{*}{HotpotQA} & 32.00          & 65.66          & 64.72          & 43.00           \\
Rewriter                        &                           & 30.00          & 62.94          & 62.99          & 40.00           \\
Rewriter+                       &                           & 27.00          & 62.61          & 65.12          & 44.00           \\
Rewriter+\&Filter &                           & \textbf{36.00} & \textbf{67.33} & \textbf{65.81} & \textbf{50.00}  \\ 
\hline
Standard                        & \multirow{4}{*}{2WikiMQA} & 23.00          & 54.11          & 64.89          & 28.00          \\
Rewriter                        &                           & 26.00          & 55.97          & 65.83 & 29.00           \\
Rewriter+                       &                           & 21.00          & 53.81          & 64.74          & 27.00           \\
Rewriter+\&Filter &                           & \textbf{28.00} & \textbf{59.62} & \textbf{66.39}           & \textbf{31.00}  \\
\hline
\end{tabular}
\vspace{-7pt}
\end{table*}

\section{One-Round Multi-Hop Reasoning QA Task}
We further extend the validation of ERAGent framework's efficacy in addressing multi-hop logic reasoning question-answering task.

\subsection{Experimental Settings}
The experimental settings are almost the same as Section~\ref{sec: one_round_open_domain_qa} except for the datasets.

\textbf{Datasets.} We selected two benchmark datasets, namely 2WIKIMQA \cite{2wikimqa} and HotPotQA \cite{hotpotqa}. Following the prior works \cite{LLM_filter,Query_Rewriting,meta_answer}, we evaluate on 100 random examples from the development set of each dataset. 

\subsection{Results}
Experimental results on one-round multi-hop reasoning QA task are in Table~\ref{tab: one-round multi-hop reasoning}.

\textbf{The effect of the Enhanced Question Rewriter.} When comparing the performance of the Rewriter+ and Rewriter settings against the Standard setting across two logic reasoning datasets, it was observed that neither of them can steadily improved answer accuracy. In some instances, these two settings even resulted in a decrease in accuracy. Furthermore, compareing Rewriter+ with Rewriter settings revealed no performance enhancement.

\textbf{The joint effect of Knowledge Filter and Enhance Question Rewriter.} The combined Rewriter+\&Filter setting surpasses other baseline models in terms of response quality, achieving a notable enhancement in all metrics. This outcome indicates that, within complex logic reasoning question-answering contexts, the synergistic effect of these two modules remains profoundly effective in improving response accuracy.

\subsection{Analysis}
The empirical outcomes derived from our investigation illuminate that deploying question-augmenting components, within the ambit of logical reasoning-oriented question-answering task, fails to consistently contribute to improve the accuracy of the responses generated of RAG. This observation underscores the inherent challenge in singularly leveraging question rephrasing modules to achieve notable improvements in answer precision within complex multi-hop reasoning tasks. Conversely, the synergistic interplay between the Enhanced Question Rewriter and the Knowledge Filter emerges as a potent catalyst for enhancing the capability of LLM  to generate more accurate responses. The strong synergistic effect between these two components is verified again.

\section{Multi-Session Multi-Round QA}
In the preceding sections, we have scrutinized the isolated/joint effect of the Enhanced Question Rewriter and Knowledge Filter within the ERAGent framework in improving response accuracy. In this section, we expand our evaluation to more closely mirror real-world application scenarios through a simulated multi-session, multi-turn conversation dataset that comprises multiple users with differing preferences. Our objective is to evaluate the capability of ERAGent to provide personalized responses to domain-specific questions for diverse users, as well as the response efficiency during the continues serving process of ERAGent.

\textbf{Research Questions.} Different from prior evaluations that only quantify the response quality in terms of accuracy for individual rounds, we adopt a more nuanced evaluation that also considers helpfulness, relevance, detailedness, personalization, consistency, and depth of the responses tailored for the interaction with individual users across multiple sessions and rounds. Additionally, we assess the response efficiency. This section aims to unravel two pivotal research questions: (3) How do ERAGent's personalized responses differentiate from non-personalized counterparts? (4) To what extent does ERAGent enhance the efficiency of knowledge retrieval? Does this enhancement compromise the response quality?

\subsection{Experimental Settings}
\textbf{Datasets.} Our exploration involved datasets pertinent to open-domain chatbots designed for tasks involving multiple consecutive sessions. Among these, the Conversation Chronicles dataset \cite{Conversation_Chronicles}, a comprehensive 1M multi-session dialogue collection, facilitates long-term conversational setups incorporating time intervals and detailed speaker relationships. However, this dataset lacks the knowledge-intensive dialogue essential for our RAG focused research. Besides, the MTbench dataset \cite{MTbench}, explicitly curated for evaluating open-domain chatbots, emphasizes consistency and depth in multi-turn dialogues, simulating user interactions with AI chatbots for inquiries and task directives across open domains. Nevertheless, MTbench's limitation to just two turns per dialogue, without multi-session interactions or diverse user data, presented significant constraints. Addressing the deficiencies of these cutting-edge datasets, we amalgamated their strengths, drawing on their design principles. We utilized the GPT-3.5 model to create the Multi-Session Multi-Round Question Answering (MSMTQA) dataset, which is designed to simulate interactions between 12 unique users, each possessing distinct personalities, and an AI chatbot across multiple sessions. These sessions are characterized by several rounds of open-domain, knowledge-intensive dialogues. The dataset encompasses conversations spanning 13 diverse themes, facilitating a comprehensive analysis of AI chatbot performance in varied interaction contexts. The nuances pertaining to the MSMTQA dataset are detailed in Appendix~\ref{sec:appendix_b}.

\begin{figure*}[!t] 
  \centering
    \includegraphics[width=6.3 in]{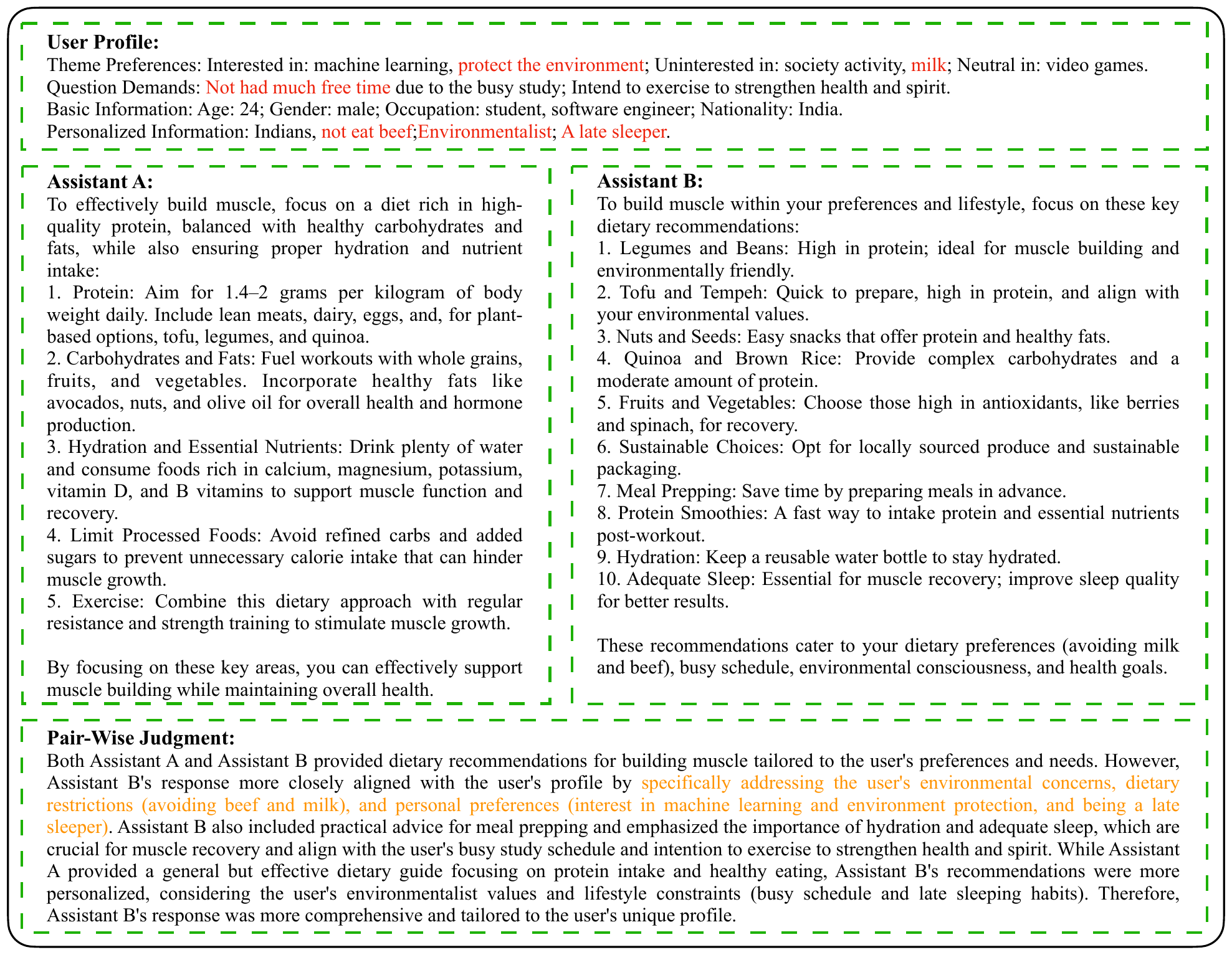}
    \vspace{-3mm}
   \caption{Two AI assistants: ERAGent without User Profile (Assistant A) and ERAGent with User Profile (Assistant B) response to a user who ask "Give me a dietary recommendation for building muscle". The User Profile is summarized from historical conversational sessions. GPT-4 is then presented with the context to determine which assistant answers better.}
    \label{fig: demo_pairwise}
\end{figure*}

\begin{figure*}[!h] 
  \centering
    \includegraphics[width=6.2 in]{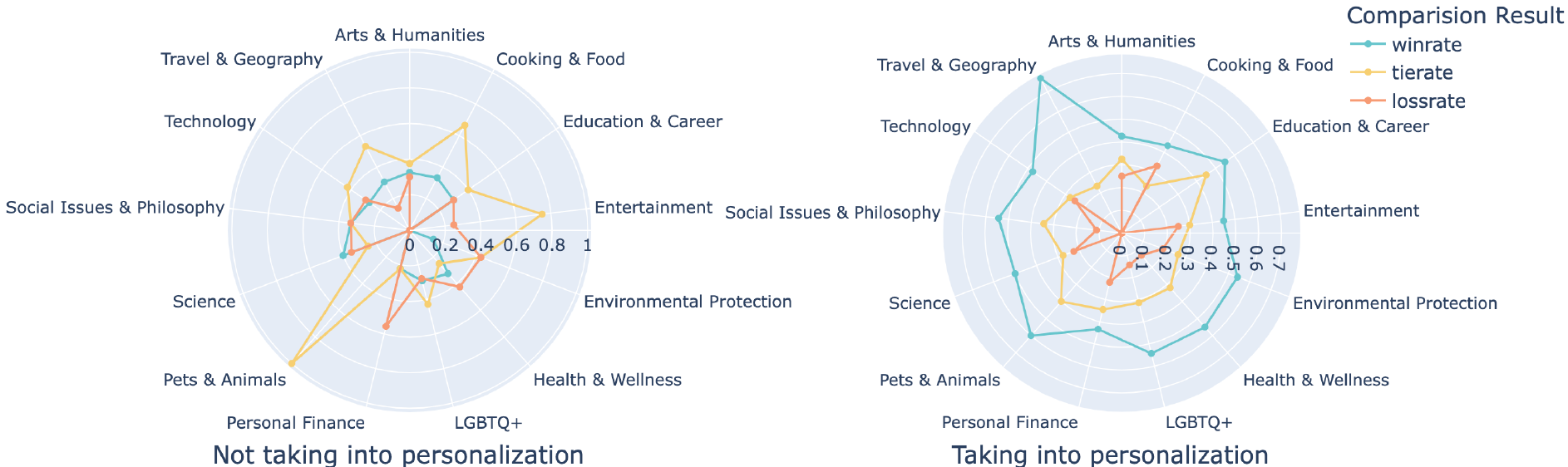}
    \vspace{-3mm}
   \caption{The results of pairwise comparisons between Assistant B and Assistant A's responses across all categories on the MSMTQA dataset.}
    \label{fig: pair_result}
\end{figure*}

\textbf{Baselines \& Evaluation Method.} 
In addressing Research Question (3), we conduct a comparative analysis between responses generated by ERAGent with Personalized LLM Reader (\textit{denoted as Assistant B}) and responses generated by ERAGent with basic LLM Reader (\textit{denoted as Assistant A}). Our evaluation methodology leverages insights from prior research on using LLM-as-a-judge \cite{MTbench} method, which demonstrated that employing GPT-4 as an adjudicator aligns with human assessments at an agreement rate exceeding 80\%. Thus, we adopt a pairwise comparison approach to discern the differences between the two sets of responses, and provide a judgment within win, loss, and tie. The dimension of this evaluative endeavor focuses on scrutinizing the aspects of helpfulness, relevance, detailedness, consistency, and depth encapsulated within the responses (\textit{Not taking into personalization}). Concurrently, the secondary facet of our comparative analysis additionally emphasizes the individuality (\textit{Taking into personalization}).

For Research Question (4), following the finish of first-time-responses to User 0's questions under the \textit{Assistant B} setting, we first clean the learned User Profile, and then adjust the popularity threshold $\tau$ of Retrieval Trigger module values to 0.2, 0.4, 0.6, and 0.8, respectively, and separately conduct a second-time-responses. The baseline first-time responses can be regarded as setting the threshold $\tau$ to 1.0 since no memory knowledge will be utilized at this value. Under the evaluation setting of \textit{Taking into personalization}, we then assess the quality of second-time-responses for pairwise comparison with the first-time responses. We record the proportions of outcomes that the-second-time response win the first-time-response as metric Win Rate. Similarly, we also record the Loss Rate and Tie Rate. Furthermore, for offering deep insights into the response dynamics and effectiveness of the ERAGent framework, we record the average time expended for a single round of conversation (Time Cost), the average count of external knowledge (External Knowledge), the average number of memory knowledge (Memory Knowledge), and the average count of knowledge that is filtered as irrelevant (Irrelevant Knowledge).

\subsection{Results}

\begin{figure*}[!h] 
  \centering
    \includegraphics[width=6.3 in]{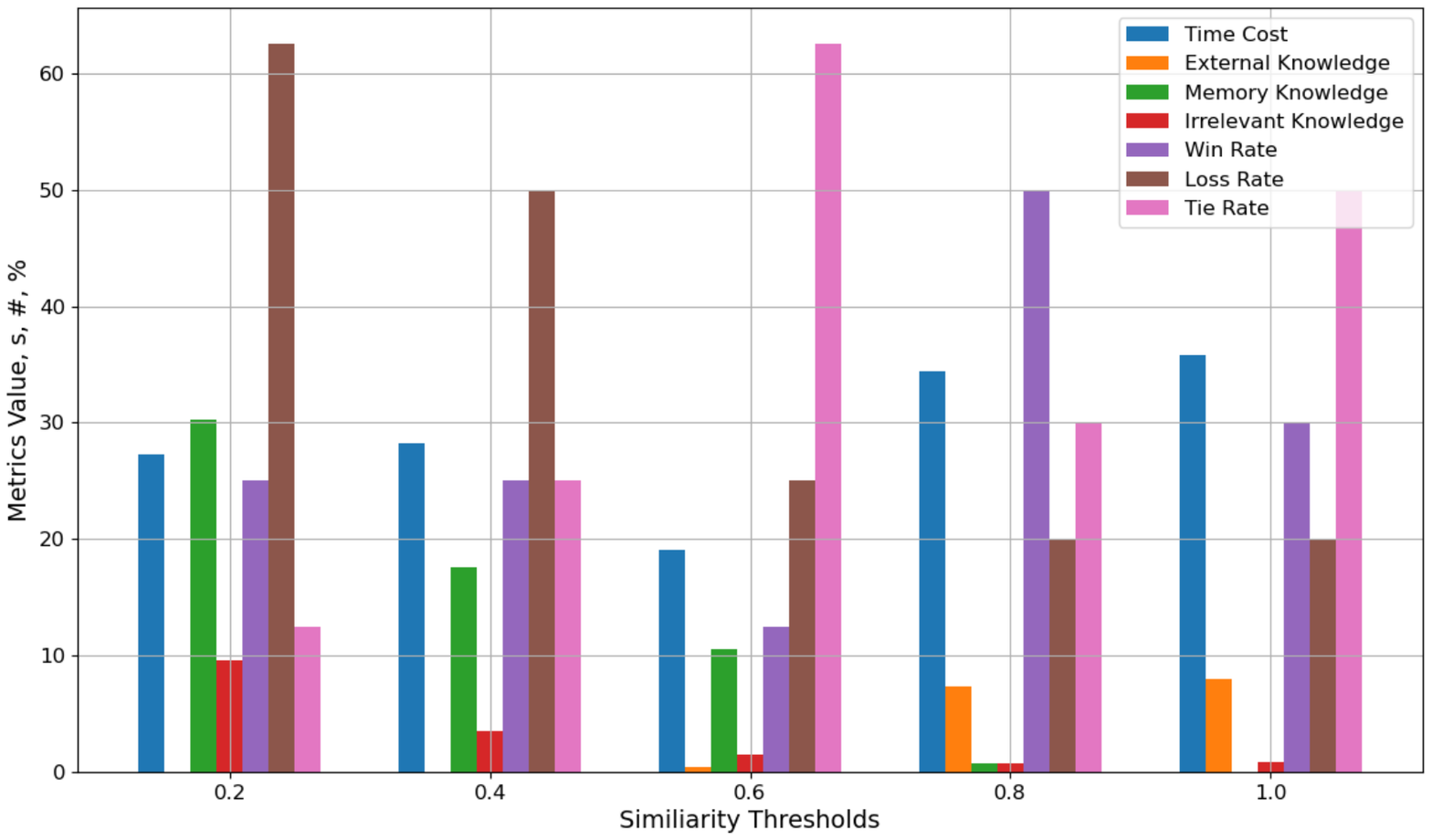}
    \vspace{-3mm}
   \caption{Metrics about response efficiency and quality alongside with the similarity threshold $\tau$.}
    \label{fig: performance_metrics}
\end{figure*}

\textbf{Pair-wise Evaluation of Response Quality}.
A case study of the pair-wise judgement result is  illustrated in Figure~\ref{fig: demo_pairwise}. This investigation elucidates the comparative analysis of response between two AI assistants, within a single-round dialogue with a user. The focal point of this case study hinges on the assessment of responses provided by Assistant A and Assistant B in the context of their consideration of the user profile. The judgement result unveils that while both assistants delivered accurate and comprehensive response. But Assistant B's response were distinguished by the personalized nature, tailored specifically to the user's preferences and needs. 

Further, the whole dataset's comparative assessment outcomes—categorized as wins, ties, and losses—in different topics are statistically summarized and visualized via a radar chart in Figure~\ref{fig: pair_result}. The left subplot of the figure exclusively concentrates on criteria including accuracy, relevance, and coherence, without taking into account the alignment with user preferences. Subsequently, the evaluation expanded to include the degree of personalization by assessing how well the responses aligned with the user's profile, with the results depicted in the figure's right subplot. It was observed that Assistant A and Assistant B often reached a draw across various topics, with instances of unilateral victories or defeats being comparatively rare. However, when takinging into personalization, Assistant B consistently outperformed Assistant A, demonstrating a superior ability to deliver tailored responses that align more closely with the user's unique profile.

\textbf{Efficiency and Quality}. The analysis of ERAGent's quality and efficiency in answering historically similar questions across different $\tau$ settings is documented in the histograms of Figure~\ref{fig: performance_metrics}, encompassing key performance metrics such as Win Rate, Loss Rate, Tie Rate, along with efficiency indicators like Time Cost, External Knowledge and so on. 

A noteworthy finding derived from the analysis of the figure is that the Time Cost, associated with the similarity threshold $\tau=0.6$, attains minimum. This is accompanied by the metric External Knowledge being nearly zero, suggesting that this configuration predominantly leverages experiential knowledge instead of external sources for the responses generation, thereby enhancing the responses efficiency. Moreover, the metric Irrelevant Knowledge observed to be low, demonstrating that this setting is capable of preventing the incorporation of extraneous content, thereby ensuring a high quality of retrieval. Remarkably, at this $\tau=0.6$, over 75\% of the responses maintained or surpassed the quality of first-time responses. Even at a $\tau$ setting of 1.0, where responses rely entirely on external knowledge, the responses are 80\% equaling or exceeding the first-time responses. 
This suggests that deploying the Experiential Learner module can achieve a significant reduction in response time—by approximately 40\%—without compromising the quality of answers. Furthermore, adjusting the threshold to 0.8 enhances the ERAGent's response quality beyond that at $\tau=1.0$, underscoring the value of leveraging highly relevant historical experience to generate superior quality responses.

\subsection{Analysis}
\textbf{Personalization in Response Generation.}
The empirical evidence presented in the case study and statistical analysis unequivocally demonstrates ERAGent's proficiency in generating responses that are not just accurate and relevant but distinctly personalized. This personalization transcends mere surface-level customization, delving into the intricacies of individual user profiles to craft responses that resonate on a more profound, personal level. The comparative analysis between Assistant A (without user profile consideration) and Assistant B (with user profile consideration) illustrates the transformative impact of personalization. Particularly noteworthy is Assistant B's consistent outperformance when evaluated against the criteria of personalization, showcasing an ability to align responses with user preferences and historical interactions. This finding substantiates the hypothesis that integrating Personalized LLM Reader module rather than basic LLM Reader module into RAG framework can significantly enhances the user experience, fostering a sense of understanding and connection between the user and the AI assistant.

\textbf{Efficiency Versus Quality in Knowledge Retrieval.}
The efficiency of ERAGent's generating responses, as gauged through various $\tau$ settings, offers fascinating insights into the delicate balance between speed and depth of information processing. The analysis reveals an optimal threshold ($\tau=0.6$) where the time cost is minimized without a commensurate dip in response quality. This configuration underscores a pivotal aspect of the ERAGent framework - its ability to leverage experiential knowledge effectively, thereby reducing reliance on external knowledge sources. This not only accelerates the response time but also minimizes the inclusion of irrelevant information, ensuring the delivery of concise, pertinent responses. Such efficiency does not undermine the quality of responses; rather, it highlights the framework's adeptness at synthesizing historical and contextual knowledge to maintain, if not enhance, the caliber of interactions.

\section{Conclusion}
In conclusion, our ERAGent framework makes substantial strides in refining RAG through innovations in enhancing question rewriting, improving response robustness with knowledge filter, and propose pioneering experiential learner to support optimizing information retrieval efficiency and improving the personalization of RAG responses. Our comprehensive evaluation demonstrates its enhanced performance across diverse datasets and tasks. While these contributions represent incremental progress rather than a paradigm shift, they underscore our commitment to advancing the field of AI with practical, scalable solutions. This research establishes a robust groundwork for subsequent explorations and enhancements within the realm of retrieval-based knowledge question-answering systems.

% \section*{Acknowledgements}
% This document has been adapted by Jordan Boyd-Graber, Naoaki Okazaki, Anna Rogers from the style files used for earlier ACL, EMNLP and NAACL proceedings, including those for
% EACL 2023 by Isabelle Augenstein and Andreas Vlachos,
% EMNLP 2022 by Yue Zhang, Ryan Cotterell and Lea Frermann,
% ACL 2020 by Steven Bethard, Ryan Cotterell and Rui Yan,
% ACL 2019 by Douwe Kiela and Ivan Vuli\'{c},
% NAACL 2019 by Stephanie Lukin and Alla Roskovskaya, 
% ACL 2018 by Shay Cohen, Kevin Gimpel, and Wei Lu, 
% NAACL 2018 by Margaret Mitchell and Stephanie Lukin,
% Bib\TeX{} suggestions for (NA)ACL 2017/2018 from Jason Eisner,
% ACL 2017 by Dan Gildea and Min-Yen Kan, NAACL 2017 by Margaret Mitchell, 
% ACL 2012 by Maggie Li and Michael White, 
% ACL 2010 by Jing-Shin Chang and Philipp Koehn, 
% ACL 2008 by Johanna D. Moore, Simone Teufel, James Allan, and Sadaoki Furui, 
% ACL 2005 by Hwee Tou Ng and Kemal Oflazer, 
% ACL 2002 by Eugene Charniak and Dekang Lin, 
% and earlier ACL and EACL formats written by several people, including
% John Chen, Henry S. Thompson and Donald Walker.
% Additional elements were taken from the formatting instructions of the \emph{International Joint Conference on Artificial Intelligence} and the \emph{Conference on Computer Vision and Pattern Recognition}.

\bibliography{custom}

\begin{thebibliography}{39}
\expandafter\ifx\csname natexlab\endcsname\relax\def\natexlab#1{#1}\fi

\bibitem[{An et~al.(2023)An, Gong, Zhong, Zhao, Li, Zhang, Kong, and
  Qiu}]{L_Eval}
Chenxin An, Shansan Gong, Ming Zhong, Xingjian Zhao, Mukai Li, Jun Zhang,
  Lingpeng Kong, and Xipeng Qiu. 2023.
\newblock \href {http://arxiv.org/abs/2307.11088} {L-eval: Instituting
  standardized evaluation for long context language models}.

\bibitem[{Borgeaud et~al.(2022)Borgeaud, Mensch, Hoffmann, Cai, Rutherford,
  Millican, Van Den~Driessche, Lespiau, Damoc, Clark
  et~al.}]{improve_by_retrieve}
Sebastian Borgeaud, Arthur Mensch, Jordan Hoffmann, Trevor Cai, Eliza
  Rutherford, Katie Millican, George~Bm Van Den~Driessche, Jean-Baptiste
  Lespiau, Bogdan Damoc, Aidan Clark, et~al. 2022.
\newblock Improving language models by retrieving from trillions of tokens.
\newblock In \emph{International conference on machine learning}, pages
  2206--2240. PMLR.

\bibitem[{Brown et~al.(2020)Brown, Mann, Ryder, Subbiah, Kaplan, Dhariwal,
  Neelakantan, Shyam, Sastry, Askell, Agarwal, Herbert-Voss, Krueger, Henighan,
  Child, Ramesh, Ziegler, Wu, Winter, Hesse, Chen, Sigler, Litwin, Gray, Chess,
  Clark, Berner, McCandlish, Radford, Sutskever, and Amodei}]{zero_shot_llm}
Tom~B. Brown, Benjamin Mann, Nick Ryder, Melanie Subbiah, Jared Kaplan,
  Prafulla Dhariwal, Arvind Neelakantan, Pranav Shyam, Girish Sastry, Amanda
  Askell, Sandhini Agarwal, Ariel Herbert-Voss, Gretchen Krueger, Tom Henighan,
  Rewon Child, Aditya Ramesh, Daniel~M. Ziegler, Jeffrey Wu, Clemens Winter,
  Christopher Hesse, Mark Chen, Eric Sigler, Mateusz Litwin, Scott Gray,
  Benjamin Chess, Jack Clark, Christopher Berner, Sam McCandlish, Alec Radford,
  Ilya Sutskever, and Dario Amodei. 2020.
\newblock \href {http://arxiv.org/abs/2005.14165} {Language models are few-shot
  learners}.

\bibitem[{Chen et~al.(2017)Chen, Fisch, Weston, and Bordes}]{DrQA}
Danqi Chen, Adam Fisch, Jason Weston, and Antoine Bordes. 2017.
\newblock Reading wikipedia to answer open-domain questions.
\newblock \emph{arXiv preprint arXiv:1704.00051}.

\bibitem[{Chowdhery et~al.(2022)Chowdhery, Narang, Devlin, Bosma, Mishra,
  Roberts, Barham, Chung, Sutton, Gehrmann, Schuh, Shi, Tsvyashchenko, Maynez,
  Rao, Barnes, Tay, Shazeer, Prabhakaran, Reif, Du, Hutchinson, Pope, Bradbury,
  Austin, Isard, Gur-Ari, Yin, Duke, Levskaya, Ghemawat, Dev, Michalewski,
  Garcia, Misra, Robinson, Fedus, Zhou, Ippolito, Luan, Lim, Zoph, Spiridonov,
  Sepassi, Dohan, Agrawal, Omernick, Dai, Pillai, Pellat, Lewkowycz, Moreira,
  Child, Polozov, Lee, Zhou, Wang, Saeta, Diaz, Firat, Catasta, Wei,
  Meier-Hellstern, Eck, Dean, Petrov, and Fiedel}]{PaLM}
Aakanksha Chowdhery, Sharan Narang, Jacob Devlin, Maarten Bosma, Gaurav Mishra,
  Adam Roberts, Paul Barham, Hyung~Won Chung, Charles Sutton, Sebastian
  Gehrmann, Parker Schuh, Kensen Shi, Sasha Tsvyashchenko, Joshua Maynez,
  Abhishek Rao, Parker Barnes, Yi~Tay, Noam Shazeer, Vinodkumar Prabhakaran,
  Emily Reif, Nan Du, Ben Hutchinson, Reiner Pope, James Bradbury, Jacob
  Austin, Michael Isard, Guy Gur-Ari, Pengcheng Yin, Toju Duke, Anselm
  Levskaya, Sanjay Ghemawat, Sunipa Dev, Henryk Michalewski, Xavier Garcia,
  Vedant Misra, Kevin Robinson, Liam Fedus, Denny Zhou, Daphne Ippolito, David
  Luan, Hyeontaek Lim, Barret Zoph, Alexander Spiridonov, Ryan Sepassi, David
  Dohan, Shivani Agrawal, Mark Omernick, Andrew~M. Dai,
  Thanumalayan~Sankaranarayana Pillai, Marie Pellat, Aitor Lewkowycz, Erica
  Moreira, Rewon Child, Oleksandr Polozov, Katherine Lee, Zongwei Zhou, Xuezhi
  Wang, Brennan Saeta, Mark Diaz, Orhan Firat, Michele Catasta, Jason Wei,
  Kathy Meier-Hellstern, Douglas Eck, Jeff Dean, Slav Petrov, and Noah Fiedel.
  2022.
\newblock \href {http://arxiv.org/abs/2204.02311} {Palm: Scaling language
  modeling with pathways}.

\bibitem[{Du et~al.(2023)Du, He, Zou, Tao, and Hu}]{shortcut}
Mengnan Du, Fengxiang He, Na~Zou, Dacheng Tao, and Xia Hu. 2023.
\newblock \href {http://arxiv.org/abs/2208.11857} {Shortcut learning of large
  language models in natural language understanding}.

\bibitem[{Gao et~al.(2023)Gao, Dai, Pasupat, Chen, Chaganty, Fan, Zhao, Lao,
  Lee, Juan, and Guu}]{RARR}
Luyu Gao, Zhuyun Dai, Panupong Pasupat, Anthony Chen, Arun~Tejasvi Chaganty,
  Yicheng Fan, Vincent Zhao, Ni~Lao, Hongrae Lee, Da-Cheng Juan, and Kelvin
  Guu. 2023.
\newblock \href {https://doi.org/10.18653/v1/2023.acl-long.910} {{RARR}:
  Researching and revising what language models say, using language models}.
\newblock In \emph{Proceedings of the 61st Annual Meeting of the Association
  for Computational Linguistics (Volume 1: Long Papers)}, pages 16477--16508,
  Toronto, Canada. Association for Computational Linguistics.

\bibitem[{Gao et~al.(2024)Gao, Xiong, Gao, Jia, Pan, Bi, Dai, Sun, Guo, Wang,
  and Wang}]{modular_rag_survey}
Yunfan Gao, Yun Xiong, Xinyu Gao, Kangxiang Jia, Jinliu Pan, Yuxi Bi, Yi~Dai,
  Jiawei Sun, Qianyu Guo, Meng Wang, and Haofen Wang. 2024.
\newblock \href {http://arxiv.org/abs/2312.10997} {Retrieval-augmented
  generation for large language models: A survey}.

\bibitem[{Gu et~al.(2023)Gu, Fan, Tang, Zhang, Zhang, Chen, Cao, Li, Madden,
  and Du}]{zeroshot_NL2SQL}
Zihui Gu, Ju~Fan, Nan Tang, Songyue Zhang, Yuxin Zhang, Zui Chen, Lei Cao,
  Guoliang Li, Sam Madden, and Xiaoyong Du. 2023.
\newblock \href {http://arxiv.org/abs/2306.08891} {Interleaving pre-trained
  language models and large language models for zero-shot nl2sql generation}.

\bibitem[{Hu et~al.(2021)Hu, Shen, Wallis, Allen-Zhu, Li, Wang, Wang, and
  Chen}]{lora}
Edward~J. Hu, Yelong Shen, Phillip Wallis, Zeyuan Allen-Zhu, Yuanzhi Li, Shean
  Wang, Lu~Wang, and Weizhu Chen. 2021.
\newblock \href {http://arxiv.org/abs/2106.09685} {Lora: Low-rank adaptation of
  large language models}.

\bibitem[{Izacard et~al.(2022)Izacard, Lewis, Lomeli, Hosseini, Petroni,
  Schick, Dwivedi-Yu, Joulin, Riedel, and Grave}]{Atlas}
Gautier Izacard, Patrick Lewis, Maria Lomeli, Lucas Hosseini, Fabio Petroni,
  Timo Schick, Jane Dwivedi-Yu, Armand Joulin, Sebastian Riedel, and Edouard
  Grave. 2022.
\newblock Few-shot learning with retrieval augmented language models.
\newblock \emph{arXiv preprint arXiv:2208.03299}.

\bibitem[{Jang et~al.(2023)Jang, Boo, and Kim}]{Conversation_Chronicles}
Jihyoung Jang, Minseong Boo, and Hyounghun Kim. 2023.
\newblock \href {https://doi.org/10.18653/v1/2023.emnlp-main.838} {Conversation
  chronicles: Towards diverse temporal and relational dynamics in multi-session
  conversations}.
\newblock In \emph{Proceedings of the 2023 Conference on Empirical Methods in
  Natural Language Processing}, pages 13584--13606, Singapore. Association for
  Computational Linguistics.

\bibitem[{Jiang et~al.(2023)Jiang, Wu, Lin, Yang, and Qiu}]{LLMLingua}
Huiqiang Jiang, Qianhui Wu, Chin-Yew Lin, Yuqing Yang, and Lili Qiu. 2023.
\newblock \href {http://arxiv.org/abs/2310.05736} {Llmlingua: Compressing
  prompts for accelerated inference of large language models}.

\bibitem[{Karpukhin et~al.(2020)Karpukhin, Oğuz, Min, Lewis, Wu, Edunov, Chen,
  and tau Yih}]{DPR}
Vladimir Karpukhin, Barlas Oğuz, Sewon Min, Patrick Lewis, Ledell Wu, Sergey
  Edunov, Danqi Chen, and Wen tau Yih. 2020.
\newblock \href {http://arxiv.org/abs/2004.04906} {Dense passage retrieval for
  open-domain question answering}.

\bibitem[{Kim et~al.(2023)Kim, Kim, Jeon, Park, and Kang}]{TOC}
Gangwoo Kim, Sungdong Kim, Byeongguk Jeon, Joonsuk Park, and Jaewoo Kang. 2023.
\newblock \href {https://doi.org/10.18653/v1/2023.emnlp-main.63} {Tree of
  clarifications: Answering ambiguous questions with retrieval-augmented large
  language models}.
\newblock In \emph{Proceedings of the 2023 Conference on Empirical Methods in
  Natural Language Processing}, pages 996--1009, Singapore. Association for
  Computational Linguistics.

\bibitem[{Kwiatkowski et~al.(2019)Kwiatkowski, Palomaki, Redfield, Collins,
  Parikh, Alberti, Epstein, Polosukhin, Devlin, Lee, Toutanova, Jones, Kelcey,
  Chang, Dai, Uszkoreit, Le, and Petrov}]{NQ}
Tom Kwiatkowski, Jennimaria Palomaki, Olivia Redfield, Michael Collins, Ankur
  Parikh, Chris Alberti, Danielle Epstein, Illia Polosukhin, Jacob Devlin,
  Kenton Lee, Kristina Toutanova, Llion Jones, Matthew Kelcey, Ming-Wei Chang,
  Andrew~M. Dai, Jakob Uszkoreit, Quoc Le, and Slav Petrov. 2019.
\newblock \href {https://doi.org/10.1162/tacl_a_00276} {Natural questions: A
  benchmark for question answering research}.
\newblock \emph{Transactions of the Association for Computational Linguistics},
  7:452--466.

\bibitem[{Lewis et~al.(2020)Lewis, Perez, Piktus, Petroni, Karpukhin, Goyal,
  K{\"u}ttler, Lewis, Yih, Rockt{\"a}schel et~al.}]{RAG}
Patrick Lewis, Ethan Perez, Aleksandra Piktus, Fabio Petroni, Vladimir
  Karpukhin, Naman Goyal, Heinrich K{\"u}ttler, Mike Lewis, Wen-tau Yih, Tim
  Rockt{\"a}schel, et~al. 2020.
\newblock Retrieval-augmented generation for knowledge-intensive nlp tasks.
\newblock \emph{Advances in Neural Information Processing Systems},
  33:9459--9474.

\bibitem[{Li et~al.(2023)Li, Nie, and Liang}]{query_router}
Xiaoqian Li, Ercong Nie, and Sheng Liang. 2023.
\newblock \href {http://arxiv.org/abs/2311.06595} {From classification to
  generation: Insights into crosslingual retrieval augmented icl}.

\bibitem[{Liu et~al.(2023)Liu, Jin, Wang, Cheng, Dou, and Wen}]{RETA_LLM}
Jiongnan Liu, Jiajie Jin, Zihan Wang, Jiehan Cheng, Zhicheng Dou, and Ji-Rong
  Wen. 2023.
\newblock Reta-llm: A retrieval-augmented large language model toolkit.
\newblock \emph{arXiv preprint arXiv:2306.05212}.

\bibitem[{Long et~al.(2023)Long, Subburam, Lowe, Santos, Zhang, Hwang, Saduka,
  Horev, Su, Cote et~al.}]{ChatENT}
Cai Long, Deepak Subburam, Kayle Lowe, Andr{\'e}~dos Santos, Jessica Zhang,
  Sang Hwang, Neil Saduka, Yoav Horev, Tao Su, David Cote, et~al. 2023.
\newblock Chatent: Augmented large language model for expert knowledge
  retrieval in otolaryngology-head and neck surgery.
\newblock \emph{medRxiv}, pages 2023--08.

\bibitem[{Lozano et~al.(2023)Lozano, Fleming, Chiang, and Shah}]{ClinfoAI}
Alejandro Lozano, Scott~L Fleming, Chia-Chun Chiang, and Nigam Shah. 2023.
\newblock Clinfo. ai: An open-source retrieval-augmented large language model
  system for answering medical questions using scientific literature.
\newblock \emph{arXiv preprint arXiv:2310.16146}.

\bibitem[{Ma et~al.(2023)Ma, Gong, He, Zhao, and Duan}]{Query_Rewriting}
Xinbei Ma, Yeyun Gong, Pengcheng He, Hai Zhao, and Nan Duan. 2023.
\newblock \href {http://arxiv.org/abs/2305.14283} {Query rewriting for
  retrieval-augmented large language models}.

\bibitem[{Mallen et~al.(2023)Mallen, Asai, Zhong, Das, Khashabi, and
  Hajishirzi}]{when_not_trust_llm}
Alex Mallen, Akari Asai, Victor Zhong, Rajarshi Das, Daniel Khashabi, and
  Hannaneh Hajishirzi. 2023.
\newblock When not to trust language models: Investigating effectiveness of
  parametric and non-parametric memories.
\newblock In \emph{Proceedings of the 61st Annual Meeting of the Association
  for Computational Linguistics (Volume 1: Long Papers)}, pages 9802--9822.

\bibitem[{Min et~al.(2020)Min, Michael, Hajishirzi, and Zettlemoyer}]{AmbigQA}
Sewon Min, Julian Michael, Hannaneh Hajishirzi, and Luke Zettlemoyer. 2020.
\newblock \href {https://doi.org/10.18653/v1/2020.emnlp-main.466} {{A}mbig{QA}:
  Answering ambiguous open-domain questions}.
\newblock In \emph{Proceedings of the 2020 Conference on Empirical Methods in
  Natural Language Processing (EMNLP)}, pages 5783--5797, Online. Association
  for Computational Linguistics.

\bibitem[{Penedo et~al.(2023)Penedo, Malartic, Hesslow, Cojocaru, Cappelli,
  Alobeidli, Pannier, Almazrouei, and Launay}]{Falcon}
Guilherme Penedo, Quentin Malartic, Daniel Hesslow, Ruxandra Cojocaru,
  Alessandro Cappelli, Hamza Alobeidli, Baptiste Pannier, Ebtesam Almazrouei,
  and Julien Launay. 2023.
\newblock \href {http://arxiv.org/abs/2306.01116} {The refinedweb dataset for
  falcon llm: Outperforming curated corpora with web data, and web data only}.

\bibitem[{Ram et~al.(2023)Ram, Levine, Dalmedigos, Muhlgay, Shashua,
  Leyton-Brown, and Shoham}]{IC-RALM}
Ori Ram, Yoav Levine, Itay Dalmedigos, Dor Muhlgay, Amnon Shashua, Kevin
  Leyton-Brown, and Yoav Shoham. 2023.
\newblock \href {http://arxiv.org/abs/2302.00083} {In-context
  retrieval-augmented language models}.

\bibitem[{R{\"o}ttger and Pierrehumbert(2021)}]{temporal_adaptation}
Paul R{\"o}ttger and Janet Pierrehumbert. 2021.
\newblock \href {https://doi.org/10.18653/v1/2021.findings-emnlp.206} {Temporal
  adaptation of {BERT} and performance on downstream document classification:
  Insights from social media}.
\newblock In \emph{Findings of the Association for Computational Linguistics:
  EMNLP 2021}, pages 2400--2412, Punta Cana, Dominican Republic. Association
  for Computational Linguistics.

\bibitem[{Savelka et~al.(2023)Savelka, Ashley, Gray, Westermann, and
  Xu}]{RAG_law_concepts}
Jaromir Savelka, Kevin~D Ashley, Morgan~A Gray, Hannes Westermann, and Huihui
  Xu. 2023.
\newblock Explaining legal concepts with augmented large language models
  (gpt-4).
\newblock \emph{arXiv preprint arXiv:2306.09525}.

\bibitem[{Shao et~al.(2023)Shao, Gong, Shen, Huang, Duan, and
  Chen}]{ITER-RETGEN}
Zhihong Shao, Yeyun Gong, Yelong Shen, Minlie Huang, Nan Duan, and Weizhu Chen.
  2023.
\newblock \href {https://doi.org/10.18653/v1/2023.findings-emnlp.620}
  {Enhancing retrieval-augmented large language models with iterative
  retrieval-generation synergy}.
\newblock In \emph{Findings of the Association for Computational Linguistics:
  EMNLP 2023}, pages 9248--9274, Singapore. Association for Computational
  Linguistics.

\bibitem[{Shi et~al.(2023)Shi, Min, Yasunaga, Seo, James, Lewis, Zettlemoyer,
  and tau Yih}]{REPLUG}
Weijia Shi, Sewon Min, Michihiro Yasunaga, Minjoon Seo, Rich James, Mike Lewis,
  Luke Zettlemoyer, and Wen tau Yih. 2023.
\newblock \href {http://arxiv.org/abs/2301.12652} {Replug: Retrieval-augmented
  black-box language models}.

\bibitem[{Wang et~al.(2023)Wang, Ma, and Chen}]{LLM_AMT}
Yubo Wang, Xueguang Ma, and Wenhu Chen. 2023.
\newblock Augmenting black-box llms with medical textbooks for clinical
  question answering.
\newblock \emph{arXiv preprint arXiv:2309.02233}.

\bibitem[{Welbl et~al.(2018)Welbl, Stenetorp, and Riedel}]{2wikimqa}
Johannes Welbl, Pontus Stenetorp, and Sebastian Riedel. 2018.
\newblock Constructing datasets for multi-hop reading comprehension across
  documents.
\newblock \emph{Transactions of the Association for Computational Linguistics},
  6:287--302.

\bibitem[{Yang et~al.(2018)Yang, Qi, Zhang, Bengio, Cohen, Salakhutdinov, and
  Manning}]{hotpotqa}
Zhilin Yang, Peng Qi, Saizheng Zhang, Yoshua Bengio, William Cohen, Ruslan
  Salakhutdinov, and Christopher~D. Manning. 2018.
\newblock \href {https://doi.org/10.18653/v1/D18-1259} {{H}otpot{QA}: A dataset
  for diverse, explainable multi-hop question answering}.
\newblock In \emph{Proceedings of the 2018 Conference on Empirical Methods in
  Natural Language Processing}, pages 2369--2380, Brussels, Belgium.
  Association for Computational Linguistics.

\bibitem[{Yao et~al.(2023)Yao, Zhao, Yu, Du, Shafran, Narasimhan, and
  Cao}]{ReAct}
Shunyu Yao, Jeffrey Zhao, Dian Yu, Nan Du, Izhak Shafran, Karthik Narasimhan,
  and Yuan Cao. 2023.
\newblock \href {http://arxiv.org/abs/2210.03629} {React: Synergizing reasoning
  and acting in language models}.

\bibitem[{Yoran et~al.(2023{\natexlab{a}})Yoran, Wolfson, Bogin, Katz, Deutch,
  and Berant}]{meta_answer}
Ori Yoran, Tomer Wolfson, Ben Bogin, Uri Katz, Daniel Deutch, and Jonathan
  Berant. 2023{\natexlab{a}}.
\newblock \href {https://doi.org/10.18653/v1/2023.emnlp-main.364} {Answering
  questions by meta-reasoning over multiple chains of thought}.
\newblock In \emph{Proceedings of the 2023 Conference on Empirical Methods in
  Natural Language Processing}, pages 5942--5966, Singapore. Association for
  Computational Linguistics.

\bibitem[{Yoran et~al.(2023{\natexlab{b}})Yoran, Wolfson, Ram, and
  Berant}]{LLM_filter}
Ori Yoran, Tomer Wolfson, Ori Ram, and Jonathan Berant. 2023{\natexlab{b}}.
\newblock \href {http://arxiv.org/abs/2310.01558} {Making retrieval-augmented
  language models robust to irrelevant context}.

\bibitem[{Zakka et~al.(2023)Zakka, Chaurasia, Shad, Dalal, Kim, Moor,
  Alexander, Ashley, Boyd, Boyd et~al.}]{Almanac}
Cyril Zakka, Akash Chaurasia, Rohan Shad, Alex~R Dalal, Jennifer~L Kim, Michael
  Moor, Kevin Alexander, Euan Ashley, Jack Boyd, Kathleen Boyd, et~al. 2023.
\newblock Almanac: Retrieval-augmented language models for clinical medicine.
\newblock \emph{Research Square}.

\bibitem[{Zhang et~al.(2024)Zhang, Zhao, Xia, Sun, Sun, Qin, Li, Zhao, Zhao,
  Cai, Zheng, Wang, and An}]{trading_agent}
Wentao Zhang, Lingxuan Zhao, Haochong Xia, Shuo Sun, Jiaze Sun, Molei Qin,
  Xinyi Li, Yuqing Zhao, Yilei Zhao, Xinyu Cai, Longtao Zheng, Xinrun Wang, and
  Bo~An. 2024.
\newblock \href {http://arxiv.org/abs/2402.18485} {A multimodal foundation
  agent for financial trading: Tool-augmented, diversified, and generalist}.

\bibitem[{Zheng et~al.(2024)Zheng, Chiang, Sheng, Zhuang, Wu, Zhuang, Lin, Li,
  Li, Xing et~al.}]{MTbench}
Lianmin Zheng, Wei-Lin Chiang, Ying Sheng, Siyuan Zhuang, Zhanghao Wu, Yonghao
  Zhuang, Zi~Lin, Zhuohan Li, Dacheng Li, Eric Xing, et~al. 2024.
\newblock Judging llm-as-a-judge with mt-bench and chatbot arena.
\newblock \emph{Advances in Neural Information Processing Systems}, 36.

\end{thebibliography}
\bibliographystyle{acl_natbib}

\appendix

\section{Appendix A}
\label{sec:appendix_a}

\begin{figure*}[t] 
  \centering
    \includegraphics[width=6.3 in]{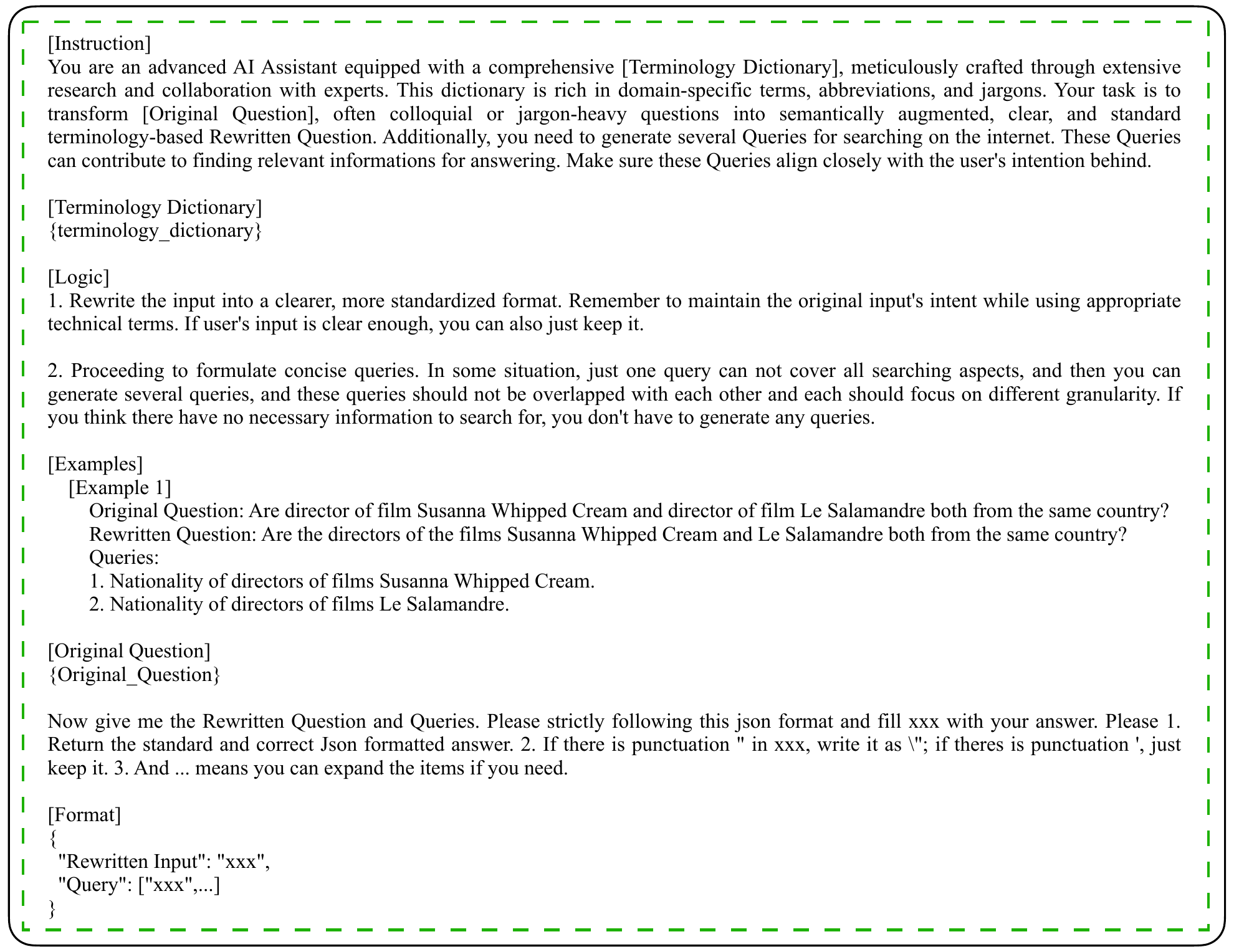}
    \vspace{-5mm}
   \caption{The default prompt for the Enhanced Question Rewriter module.}
    \label{fig: Prompt_Question_Rewriter}
\end{figure*}

\begin{figure*}[h] 
  \centering
    \includegraphics[width=6.3 in]{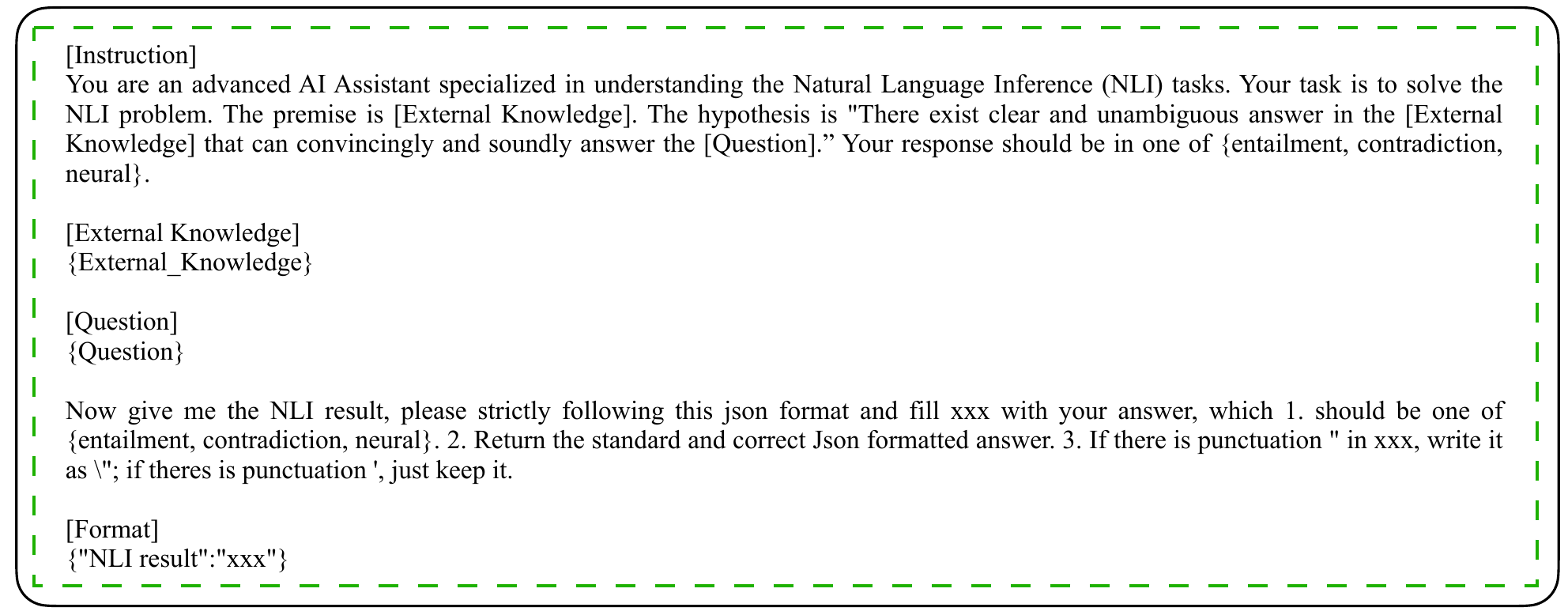}
    \vspace{-5mm}
   \caption{The default prompt for the Knowledge Filter module.}
    \label{fig: Prompt_Knowledge_Filter}
\end{figure*}

\begin{figure*}[h] 
  \centering
    \includegraphics[width=6.3 in]{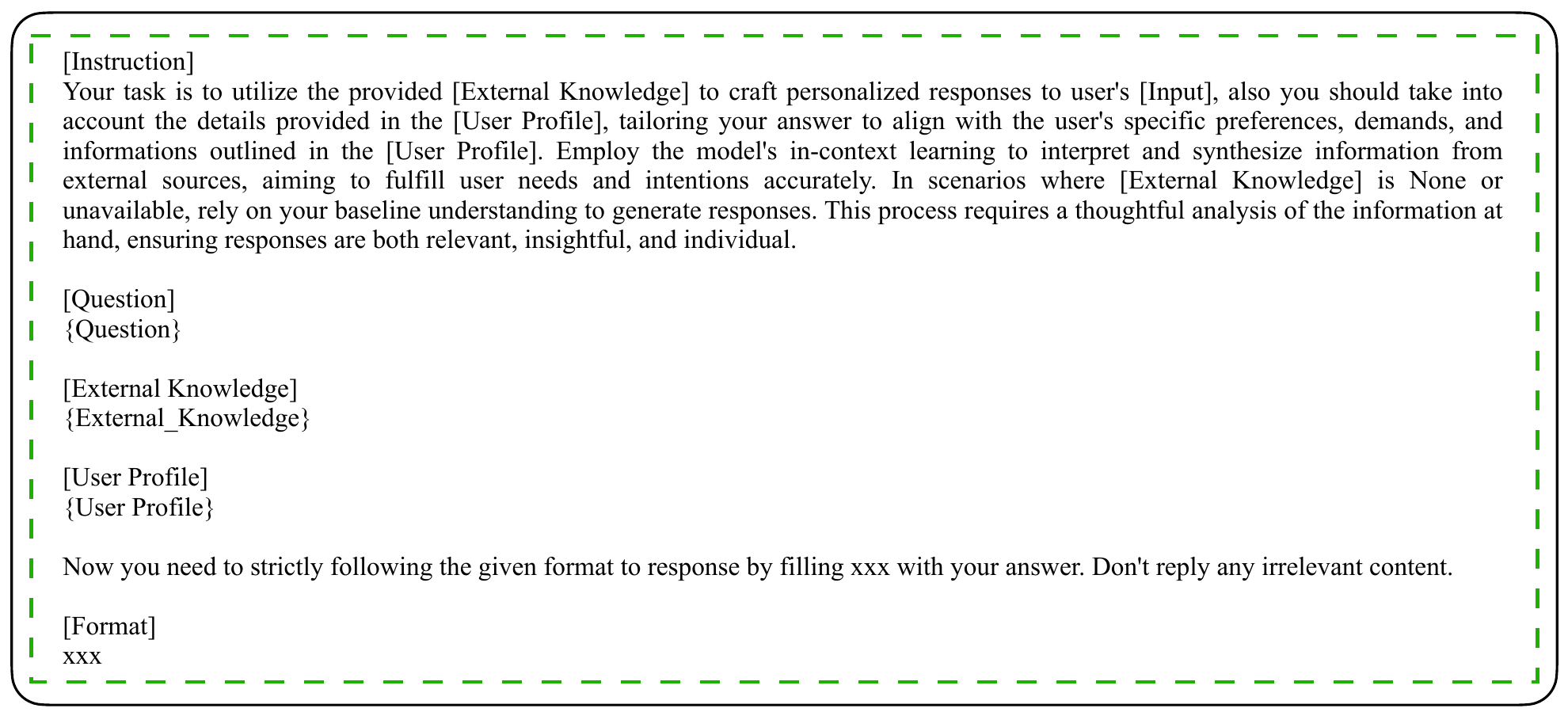}
    \vspace{-5mm}
   \caption{The default prompt for the Personalized LLM Reader module.}
    \label{fig: Prompt_Personalized_LLM_Reader}
\end{figure*}

\begin{figure*}[h] 
  \centering
    \includegraphics[width=6.3 in]{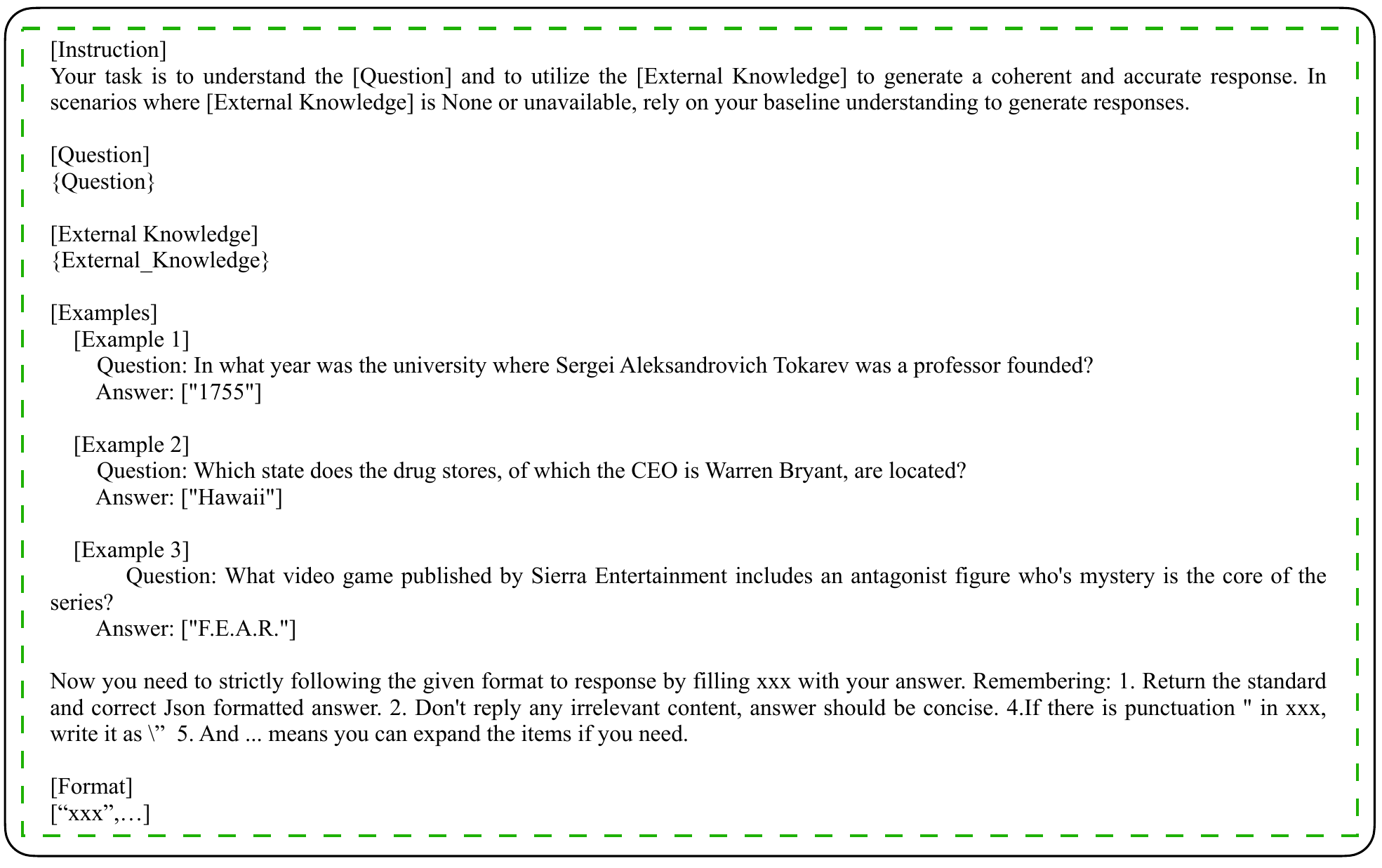}
    \vspace{-5mm}
   \caption{The default prompt for the LLM Reader module.}
    \label{fig: Prompt_LLM_Reader}
\end{figure*}

In the ERAGent framework, the prompts employed by each module are tailored to address the nuances of specific datasets and task scenarios, reflecting slight adjustments and emphases. This appendix section serves to offer exemplary instances of default prompts for illustrative purposes. For instance, the prompt template for the Enhanced Question Rewriter module is depicted in Figure~\ref{fig: Prompt_Question_Rewriter}. In a similar vein, we showcase the prompt terminology utilized in the Knowledge Filter module. This module leverages large language models (LLMs) for executing natural language inference (NLI) tasks, as delineated in Figure~\ref{fig: Prompt_Knowledge_Filter}. Moreover, the prompts employed in the Personalized LLM Reader and the LLM Reader modules are presented in Figures~\ref{fig: Prompt_Personalized_LLM_Reader} and ~\ref{fig: Prompt_LLM_Reader}, respectively.

\section{MSMTQA dataset}\label{sec:appendix_b}

\begin{figure*}[h] 
  \centering
    \includegraphics[width=6.3 in]{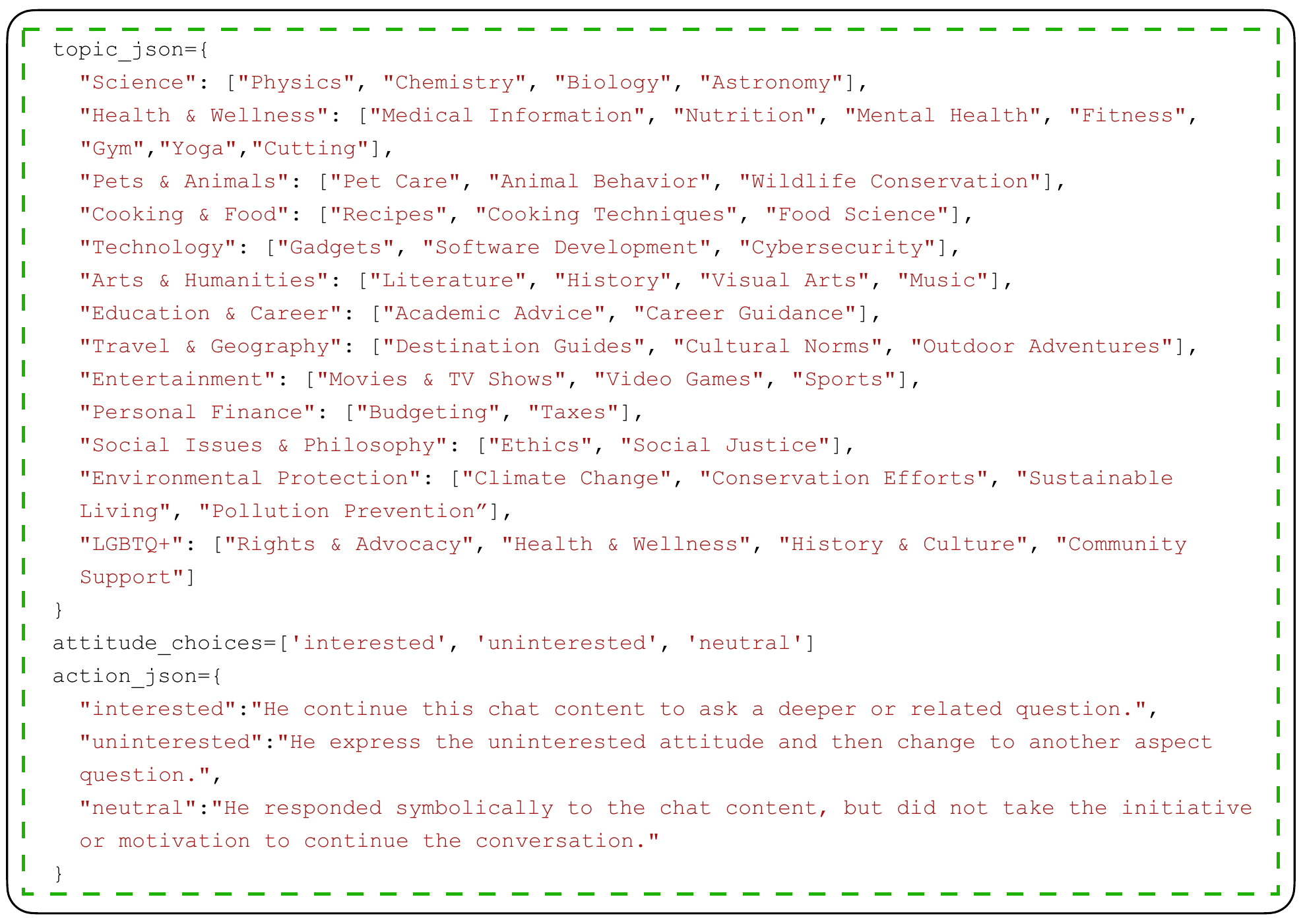}
    \vspace{-5mm}
   \caption{Thirteen main topics and subtopics of the conversational content. As well as the three types of attitude of user responses, alongside with their detailed response actions.}
    \label{fig: topic_attitude}
\end{figure*}

\begin{figure*}[h] 
  \centering
    \includegraphics[width=6.3 in]{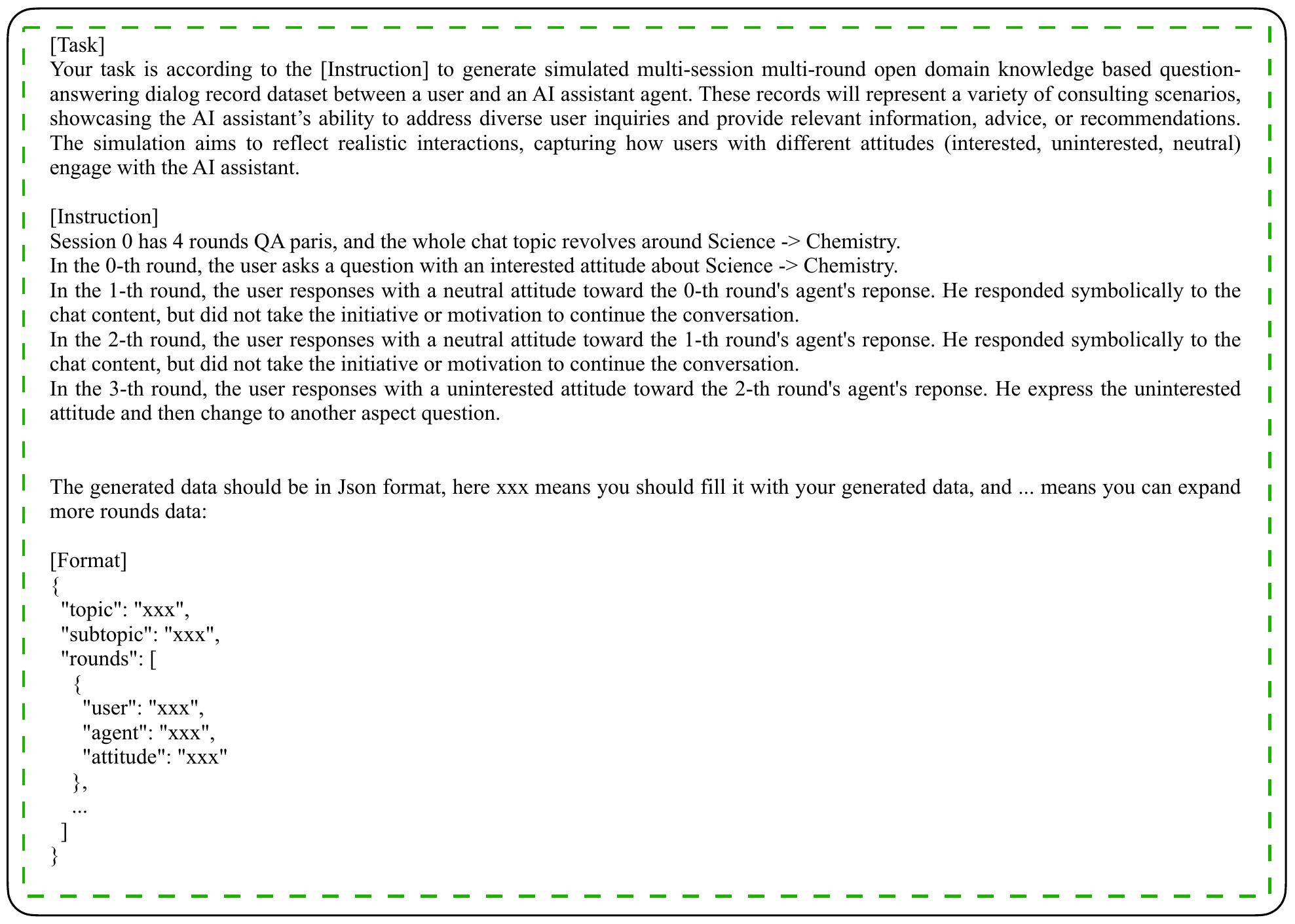}
    \vspace{-5mm}
   \caption{The prompt for the generation of simulated MSMTQA dataset.}
    \label{fig: Prompt_MSMTQA}
\end{figure*}

We initially delineate the primary and subsidiary themes within the conversational content, alongside elucidating the specifics concerning the actions correlated with the attitudes manifested in user responses, as depicted in Figure~\ref{fig: topic_attitude}. 
Subsequently, we employ the prompt template presented in Figure~\ref{fig: Prompt_MSMTQA}, systematically generating conversational data between a user and an AI assistant on a session-by-session basis. The instruction part in this prompt is controlled by a stochastic process that dictates the thematic direction of the dialogue and the user's interest action toward each response from the AI assistant. By adopting this method, we have compiled conversational data for twelve users, culminating in the formation of the MSMTQA dataset.

\end{document}